\documentclass{article}

\usepackage{PRIMEarxiv}

\usepackage[utf8]{inputenc} % allow utf-8 input
\usepackage[T1]{fontenc}    % use 8-bit T1 fonts
\usepackage{hyperref}       % hyperlinks
\usepackage{url}            % simple URL typesetting
\usepackage{booktabs}       % professional-quality tables
\usepackage{amsfonts}       % blackboard math symbols
\usepackage{nicefrac}       % compact symbols for 1/2, etc.
\usepackage{microtype}      % microtypography
\usepackage{lipsum}
\usepackage{fancyhdr}       % header
\usepackage{graphicx}       % graphics
\graphicspath{{media/}}     % organize your images and other figures under media/ folder

\usepackage{amssymb,amsmath}
\usepackage{graphicx,psfrag,epsf,subfig}
\usepackage{enumerate}
\usepackage{natbib}
\usepackage{multirow,multicol}
\usepackage{array}
\usepackage{bm}
\usepackage{float}
\usepackage{fancyhdr} 
\usepackage{framed}

\usepackage{algorithm}
\usepackage{algorithmicx}
\usepackage{algpseudocode}

\newcommand{\Mean}{{\mathbb{E}}}
\newcommand{\Var}{{\mbox{Var}}}

\newtheorem{coro}{Corollary}[section]

\newtheorem{theorem}{Theorem}  
\newtheorem{lemma}{Lemma}[section]
\newtheorem{remark}{Remark}[section]
\newtheorem{assumption}{Assumption}[section]

\usepackage{xr}

\usepackage{chngcntr}

\title{Heterogeneous Synthetic Learner for Panel Data
}

\author{
  Ye Shen \\
  North Carolina State University \\
  \texttt{yshen28@ncsu.edu} \\
   \And
  Runzhe Wan\\
   Amazon\\
  \texttt{runzhe.wan@gmail.com} \\
     \And
  Hengrui Cai\\
   University of California Irvine\\
  \texttt{hengrc1@uci.edu} \\
     \And
  Rui Song\\
   North Carolina State University \\
  \texttt{rsong@ncsu.edu} \\
}

\begin{document}
\maketitle

\begin{abstract}
In the new era of personalization, learning the heterogeneous treatment effect (HTE) becomes an inevitable trend with numerous applications. Yet, most existing HTE estimation methods focus on independently and identically distributed observations and cannot handle the non-stationarity and temporal dependency in the common panel data setting.  The treatment evaluators developed for panel data, on the other hand, typically ignore the individualized information. To fill the gap, in this paper, we initialize the study of HTE estimation in panel data. Under different assumptions for HTE identifiability, we propose the corresponding heterogeneous one-side and two-side synthetic learner, namely H1SL and H2SL, by leveraging the state-of-the-art HTE estimator for non-panel data and generalizing the synthetic control method that allows flexible data generating process. 
We establish the convergence rates of the proposed estimators. 
The superior performance of the proposed methods over existing ones is demonstrated by extensive numerical studies. 
\end{abstract}

% % keywords can be removed
% \keywords{Heterogeneous treatment effect \and Individual treatment effect \and Panel Data \and Meta-Leaners}

\section{Introduction} \label{sec:intro}

Evaluating the treatment effect from panel data has become an increasingly important problem in numerous areas including public health  \citep{cole2020impact, goodman2020using}, politics \citep{abadie2010synthetic, sabia2012effects},  economics \citep{cavallo2013catastrophic, dube2015pooling}, etc. 
During the past decades, a number of methods have been developed to estimate the average treatment effect (ATE) from panel data, including the celebrated Difference-in-Differences (DiD) \citep{abadie2005semiparametric} and the Synthetic Control (SC) method \citep{abadie2003economic,abadie2010synthetic}. 
Yet, due to the heterogeneity of individuals in response to treatments, there may not exist one single uniformly optimal treatment across individuals.  
Thus, one major focus in causal machine learning is to access the Heterogeneous Treatment Effect (HTE) \citep[see e.g.,][]{athey2015machine,shalit2017estimating,wager2018estimation,kunzel2019metalearners,farrell2021deep} that measures the causal impact within a given group. 
% , as a fundamental component in a number of existing high-profile successes including policy optimization  \citep{chakraborty2013statistical,greenewald2017action} and policy evaluation \citep{swaminathan2017off,kallus2018balanced}.  % dimakopoulou2017estimation
 %, the advertisement and marketing \citep{athey2015machine, farias2021learning}, public policy \citep{green2010modeling, stoffi2018estimating,chernozhukov2018generic,kunzel2019metalearners}, job market \citep{imai2013estimating, shalit2017estimating}, education \citep{bargagli2019heterogeneous}, etc.  
% that accesses  the benefit of adopting a new intervention across the entire population 
% \hl{Emphasize the importance of HTE: interpretability, personalized decision making, etc.}
Detecting such a heterogeneity in panel data hence becomes an inevitable trend in the new era of personalization. %with many vital applications including the advertisement and marketing \citep{athey2015machine, farias2021learning}, public policy \citep{green2010modeling, stoffi2018estimating,chernozhukov2018generic,kunzel2019metalearners}, job market \citep{imai2013estimating, shalit2017estimating}, education \citep{bargagli2019heterogeneous}, etc. 
% \hl{Consider merging with the first several sentences of this paragraph.}

However, estimating HTE in panel data is surprisingly underexplored in the literature. 
On the one hand,  despite the fact that there are many  methods for the HTE estimation 
\citep[see e.g.,][and the reference therein]{athey2016recursive,johnson2019detecting,kunzel2019metalearners,nie2021quasi}, most of these works focus on independently and identically distributed (i.i.d.) observations and thus are infeasible to handle the non-stationarity and temporal dependency in the common panel data setting. 
On the other hand, in contrast to the popularity of estimating ATE in panel data as mentioned above, limited progress has been achieved for HTE.

% and overcome the aforementioned challenges
\textbf{Contributions. }  
To fill the gap, in this paper we aim to identify treatment heterogeneity in panel data. 
Our contributions are threefold. % to the best of our knowledge, 5
First,  we pioneer the literature to propose methods for estimating heterogeneous treatment effects in panel data. % with multiple treated units. 
 To characterize the heterogeneity, we first conceptualize both HTE and the Heterogeneous Treatment Effects on the Treated (HTT) in panel data. 
 We further discuss the causal identifiability for these effects. 
 
 Second, under different ignorability and model assumptions, we propose the corresponding one-side and two-side learners, which utilize the information from control units solely and that from both control and treatment units, respectively. 
 In each case, we first propose the heterogeneous cross learner (X-Learner) for reducing the bias and then design the doubly robust (DR) version for better efficiency. 
 The four algorithms are named  Heterogeneous One-side Synthetic X-Learner (H1SXL), Heterogeneous One-side Synthetic Doubly Robust Learner (H1SDRL),  Heterogeneous Two-side Synthetic X-Learner (H2SXL), and Heterogeneous Two-side Synthetic Doubly Robust Learner (H2SDRL). 
% Specifically, when HTT is identifiable while HTE is unidentifiable, we propose Heterogeneous One-side Synthetic X-Learner (H1SXL) and Heterogeneous One-side Synthetic Doubly Robust Learner (H1SDRL). 
% In the case that HTT and HTE are both identifiable, we propose Heterogeneous Two-side Synthetic X-Learner (H2SXL) and Heterogeneous Two-side Synthetic Doubly Robust Learner (H2SDRL).
 Our proposed learners leverage state-of-the-art HTE estimators for i.i.d. data and the SC method that allows flexible data-generating processes.
%\todo{HC: Shall we mention H1SXL/DRL somewhere? We'd better break this contribution paragraph for better readability.}

Third, we establish the convergence rates of the proposed estimators.
Take the two-side learners as an example. 
With a linear HTE function, the mean squared error of  H2SXL has a rate of $O_p\left(\left( n^{-1} + m^{-1}  + T_0^{-1}  \right)  T_1^{-1}   \right) $  and  that of H2SDRL has $ O_{p} \left( \left\{(m+n)^{-1}   +T_0 ^{-1}\right\}T_1^{-1}\right)$. 
% The decay with both dimensions of the panel data clearly shows the advantage of leveraging the panel structure in HTE estimation. 
%the H1SXL method works for more general cases with {weak assumptions} required for causal identification, and achieves a convergence rate of mean square error at $O_p\left\{ {   \log (n)}/{T_{0}}   \right\}$, where $n$ is number of control units used to estimate HTE and $T_0$ corresponds to the length of the untreated period for fitting SC. With a slightly stronger identification assumption, the proposed H2SL yields a faster convergence rate when the size of control units dominates that of treated units. 
%Lastly, we use ex
Extensive simulation studies and one real application to finance have been conducted to illustrate the superior finite-sample performance of the proposed methods over existing baselines. 
% is further justified by .

% The rest of this paper is organized as follows. We first introduce the methods of estimating the counterfactual outcomes using the canonical synthetic control Method for a single treated unit and the extended Synthetic Control Method for multiple treated units in Section \ref{sec:sc}. Tn Section \ref{sec:isc}, we present our proposed Synthetic X-Learner method. In Section \ref{sec:theory}, we establish the convergence rate of our Synthetic X-Learner method and sufficient conditions for reaching this rate. We examine the finite sample properties of our method in Section \ref{sec:sim}.

\section{Related Works} \label{sec:review}
% \todorw{1. HTE estimation, but they are mostly restricted to iid data 2. Panel data estimator, where SC is popular and why 3. A paragraph on the most important or related advanced of SC 4. The two related work and our unique advantages}
\textbf{ATE estimation with panel data.} 
ATE estimation with panel data has been a long-standing topic in causal inference. Among the many proposed methods \citep{athey2021matrix}, DiD \citep{abadie2005semiparametric} and SC \citep{abadie2003economic,abadie2010synthetic,abadie2021using} are arguably the most popular. The DiD method requires strong assumptions about the common trends between the treatment units and the control units,   and hence is usually criticized as restrictive. In contrast, SC is more flexible with the data-generating process \citep{kreif2016examination}, by approximating the counterfactual outcome under control of the treated unit with the  weighted average of control units, i.e., constructing a synthetic control unit. 
%  Not sure if the below adds value to our paper
%  SC relies on the assumption that the treated and control units follow parallel trends without treatment. 
%  Relaxation of this assumption has been pioneered by \citet{ferman2019synthetic, doudchenko2016balancing,bottmer2021design} and \citet{ben2021augmented} through adjusting the restrictions on the weights and using regression-based methods. 
% To avoid extrapolation, the weights are restricted to be nonnegative and sum to one in the classic Synthetic Control method. 
%Based on this assumption
For inference,  \citet{chernozhukov2021exact} introduced an exact and robust conformal inference method based on permutation, and \citet{li2020statistical} derived the distribution theory for SC estimator using projection theory with a sub-sampling method to conduct hypothesis testing and construct confidence intervals.  
Most recently, the SC methods have also found applications in settings with multiple treated units \citep{dube2015pooling, robbins2017framework, abadie2021penalized}, which however are still restricted to the estimation of ATE. 
%See \citet{abadie2021using} for comprehensive literature.

\textbf{HTE estimation with non-panel data.} 
The past decade has witnessed increasing interest in HTE estimation. There are two main groups of HTE estimation techniques \citep{jacob2021cate}. The first group estimates HTE based on tree methods \citep{athey2016recursive, athey2019generalized, powers2018some,hahn2020bayesian}. 
The second group, known as meta-learners, breaks down the task of estimating HTE into several regression problems that can be solved using off-the-shelf machine learning methods.
We could further categorize the meta-learners into two subgroups. The first group estimates HTE by modeling the outcome directly. For example,  \citet{hansotia2002incremental} proposed  the T-learner which models `two' estimators of conditional means under the treated and control groups, and the final HTE estimator is the difference of the two conditional means.
\citet{hill2011bayesian} came up with the S-learner using a `single' regression model for conditional outcome function with the contrast being the estimated HTE function.
\citet{nie2021quasi} designed R-Learner, which estimates marginal effects and treatment propensities first and then uses a data-adaptive objective function to learn HTE non-parametrically.
The second subgroup estimates HTE based on imputed ITE or its pseudo outcome. 
 \citet{kunzel2019metalearners} combined the above mentioned T-learner with the   S-learner and developed X-learners based on cross-fitting and out-of-bag prediction to further reduce the bias.
\citet{kennedy2020optimal} studied a general class of doubly robust algorithms built on the T-learner and a modified version of augmented inverse probability weighting pseudo outcome.
However, one common limitation of these works is that they are only applicable to independently and identically distributed observations. 
% invariant over time. 

\textbf{HTE estimation with panel data.} 
% To the best of our knowledge, 
There is very limited attention paid to the estimation of HTE in panel data. 
\citet{viviano2019synthetic} proposes a boosting algorithm to estimate HTE in a time-dependent setting. 
Yet, their method focuses on the time-dependent heterogeneity in a \textit{single} treated unit and is hard to handle \textit{multiple} treated units.  \citet{nie2019nonparametric} introduces R-DiD, an HTE estimator based on DiD  under a two-way fixed effect model.
However, their method requires that observed and unobserved predictors of the outcome remain constant over time, which is not applicable to the common panel data setting where the outcome depends on time-variant latent factors as focused in our work. 
Moreover, the parallel trend assumption needed for DiD is known as restricted. 

% Given the time-dependent data with treatment effect heterogeneity, ATE estimation might be biased and can no longer well capture the treatment effects \citep{de2020two}.   

  \section{Problem Formulation} \label{sec:fram}
% We start by introducing some notations and assumptions for establishing the concept and identification of heterogeneous treatment effects in panel data. 
\subsection{Setup}
In this paper, we focus on panel datasets with multiple units and with a time series of outcomes for each unit.  
Consider an experimental or observational study with $N = m+n$ units and $T = T_0+T_1$ total time points. 
Without loss of generality, we assume that the first $m$ units are treated units and the last $n$ units are control units. 
For simplicity of notations, we also make the common assumption that all treated units begin receiving their treatments from the same time point $t = T_0 + 1$.   
% Our methodology is equally applicable when the start time differs.  (We can not deal with staggered adoption cases)
% with the $T_0$ time points before the treatment and $T_1$ time points after the treatment. 
Assume that each unit $i$ has a $d$-dimensional time-invariant feature vector $\boldsymbol{X}_{i} \in \mathcal{X} \subseteqq \mathbb{R}^{d}$ (with intercept), and receives an unit-level outcome  $Y_{i,t}$ at each time step $t$. 
%Denote $\boldsymbol{Y}_{i}^{pre} = [ Y_{i,1}, \cdots,Y_{i, T_{0}}]^{\prime}$ as all outcomes observed for unit $i$ before the experiment and $\boldsymbol{Y}_{i}^{post} = [ Y_{i,T_{0}+1}, \cdots,Y_{i, T_{0}+T_{1}}]^{\prime}$ as all outcomes observed for unit $i$ after the experiment.
%Our data structure can hence be summarized in the following matrix:
%\begin{equation*}
%\resizebox{\hsize}{!}{$\left(\begin{array}{ccc|ccc} \boldsymbol{X}_{1} & \ldots & \boldsymbol{X}_{m} & \boldsymbol{X}_{m+1} & \ldots & \boldsymbol{X}_{m+n}  \\  \hline  Y_{1,1}  & \ldots &Y_{m,1}  & Y_{m+1,1}  & \ldots & Y_{m+n, 1} \\ \vdots  &  & \vdots  & \vdots & & \vdots \\ Y_{1, T_{0}}&\ldots &Y_{m, T_{0}} &Y_{m+1, T_{0}} & \ldots &Y_{m+n, T_{0}} \\ \hline Y_{1, T_{0}+1}&\ldots & Y_{m, T_{0}+1}& Y_{m+1, T_{0}+1}& \ldots & Y_{m+n,T_{0}+1}\\ \vdots  &  & \vdots & \vdots &  & \vdots \\ Y_{1, T_{0}+T_{1}}&\ldots & Y_{m, T_{0}+T_{1}}&  Y_{m+1, T_{0}+T_{1}} & \ldots & Y_{m+n, T_{0}+T_{1}} \end{array}\right),$}
%\end{equation*}
%where each column denotes the observations for a single unit and each row denotes the observations at a time point expect the first row as the time-invariant feature information.   
Denote $D_{i,t}$ as the treatment assignment indicator, where $D_{i,t} = 1$  if the unit $i$ is treated at time $t$, and $D_{i,t} = 0$ otherwise. 
In our setting, we have $D_{i,t} = 0, t = 1, 2, \cdots, T_0$ and $D_{i,t} = 1, t = T_0+1, T_0+2, \cdots, T$ for treated units, and $D_{i,t} = 0,  t = 1, 2, \cdots, T$ for control units. We assume the treatment assignment follows the propensity function
% as the probability of receiving the treatment given features  $\boldsymbol{X}=\boldsymbol{x}$ as  
 $e(\boldsymbol{x})=\operatorname{Pr}(D_{i,T}=1 \mid \boldsymbol{X}=\boldsymbol{x})$.  Throughout this paper, we adopt the potential outcome framework in causal inference \citep{splawa1990application,rubin1974estimating}. 
Specifically, let  $Y_{i,t}(1)$ and $Y_{i,t}(0)$ be the potential outcome for unit $i$ in time point $t$ that would be observed if this unit receives the treatment or control, respectively. For any real numbers $a, b \in \mathcal{R}$, denote $a \vee b =\max\{a,b\}$.

\subsection{Causal Estimands of Interest}\label{sec:HTE}

In this paper, we are interested in measuring the heterogeneous treatment effect using panel data.
%of adopting a new treatment over the control 
  To this end, we introduce several causal estimands at the unit level and discuss their relationships as follows. 
First,  we define the Individual Treatment Effect (ITE) for the $i$-th unit at time point $t$ as $\Delta_{i,t} =Y_{i,t}(1) - Y_{i,t}(0)$. 
Assuming that these ITEs follow a stationary process over time, the Average Treatment Effect (ATE) can be defined as $\Mean(\Delta_{i,t})$ by taking the expectation over time points and units. 
% Similarly, we can  define the Average Treatment effect on the Treated (ATT) as  $\Mean (\Delta_{i,t}|D_{i,T} = 1)$. 
% in the response to the treatment
Due to the heterogeneity among individuals, a more informative quantity is needed to describe the treatment effects for units with different features. 
% , also known as the conditional average treatment effect,
This motivates the definition of the Heterogeneous Treatment Effect (HTE) as the expectation of the ITE  conditioned on the feature information $\boldsymbol{X}_i=\boldsymbol{x}$ as
\begin{equation*}\label{eqn:HTE}
    \tau(\boldsymbol{x}):=\mathbb{E}[\Delta_{i,t}\mid \boldsymbol{X}_i=\boldsymbol{x}]=\mathbb{E}[Y_{i,t}(1) - Y_{i,t}(0) \mid \boldsymbol{X}_i=\boldsymbol{x}]. 
\end{equation*}
We have that $\text{ATE} =\mathbb{E}\{\tau(\boldsymbol{x})\} $ by taking expectation over the feature distribution. 
The HTE function $\tau(\boldsymbol{x})$ characterizes the heterogeneity 
%in the group 
of units with features $\boldsymbol{x}$, and thus is more informative than ATE. 
% To capture the heterogeneity in the treated group, w
We similarly define the Heterogeneous Treatment Effects on the Treated  (HTT), also known as conditional average treatment effect on the treated, as the expectation of ITE over experiment time points given the feature information $\boldsymbol{X}_i=\boldsymbol{x}$ for the treated units:
\begin{equation*}
	\delta(\boldsymbol{x}) := \mathbb{E}[\Delta_{i,t}\mid \boldsymbol{X}_i=\boldsymbol{x},D_{i,t} = 1],
\label{eqn:HTT}
\end{equation*}
%\hl{This definition is not rigorous. LFS is $t$-free. It can be $D_{i,T}$.} 
%where $D_{i,t} = 1$ is an indicator for the  treated units  and $t> T_0$ corresponds to the experiment periods. 
Unlike the HTE, HTT focuses on the treatment heterogeneity for the treated units only.

To establish the identification of causal estimands, we make the following standard assumptions in causal inference literature \citep[see e.g.,][]{athey2016recursive,abadie2021using}.
%It is well known that the identification of causal estimands requires assumptions  
%Throughout this paper, we consider three common assumptions in the literature: the no anticipation assumption,  the stable unit treatment value assumption(SUTVA), and the conditional independence assumption. 
% These are all   in the literature to ensure the identifiability of the causal effects in panel data.
\begin{assumption}(No Anticipation)\label{ass1}
For any unit $i$ and time $t \le T_0$, we have  $Y_{i,t} = Y_{i,t}(0)$.
\end{assumption}
\begin{assumption}(Stable Unit Treatment Value Assumption)\label{SUTVA} For any unit $i$ and time $t \ge T_0+1$,  we have $Y_{i,t}=Y_{i,t}(1)  D_{i,t} + Y_{i,t}(0) (1-D_{i,t})$.
\end{assumption}
\begin{assumption}(Ignorability)\label{ass:ci} For any unit $i$ and time $t\geq T_0$, (a)  $Y_{i,t}(0) \perp D_{i,t} \mid \boldsymbol{X}_{i},\{Y_{i,j}\}_{j=1}^{T_0}$  (b) $Y_{i,t}(1) \perp D_{i,t} \mid  \boldsymbol{X}_{i},\{Y_{i,j}\}_{j=1}^{T_0}$.
\end{assumption}
Assumption \ref{ass1} states that the treatment has no effect on the outcome before the implementation period $T_0+1$. Assumption \ref{SUTVA} requires that the observed outcome of a particular unit depends only on its received treatment, without the affection of other units' treatment assignments. With Assumption \ref{ass1} and \ref{SUTVA} 
%and the definition of ITE
, we can express  the observed outcome of interest $Y_{i,t}$ for unit $i$ at time point $t$ as 
\begin{equation}\label{mid_step}
Y_{i,t} = Y_{i,t}(0) + D_{i,t}\Delta_{i,t}.
\end{equation}
Assumption \ref{ass:ci} (a) is a core assumption for treatment effect estimation with panel data. 
It requires that given all the observed information prior to the treatment, no unmeasured confounders affect both the outcome and the treatment assignment. 
With Assumption \ref{ass:ci} (a), we are able to identify the potential outcomes under control for all the treated units, which enables us to measure HTT as discussed in Section \ref{sec:Estimator_one_side}. Assumption \ref{ass:ci} (a) combined with Assumption \ref{ass:ci} (b) is a stronger ignorability assumption commonly assumed in the literature which holds under randomized studies or when all the confounders are captured in $\boldsymbol{x}$. 
Under this stronger ignorability assumption, the HTT $\delta(\boldsymbol{x})$ equals to the HTE $\tau(\boldsymbol{x})$ and both are identifiable.  
We summarize the identifiability for different estimands in Table \ref{tab:ci}. 
%\textcolor{red}{For concreteness, we focus on estimating HTT throughout the remaining paper. 
%When the stronger ignorability holds, the two estimands are equal and we can also utilize more information to gain efficiency (will be discussed in Section \ref{sec:SCX}). }

\begin{table}[t]
\caption{Causal identifiability under different assumptions.}
\label{tab:ci}
\begin{center}
\begin{tabular}{|cccc|cc|c|c|}  \hline
 \multicolumn{4}{|c|}{ Assumptions} &   \multicolumn{2}{|c|}{ Identifiability} &Causal Equivalence& Estimation \\ \hline 
  \ref{ass1}  & \ref{SUTVA} &\ref{ass:ci} (a)& \ref{ass:ci} (b)&HTT  & HTE & HTE = HTT & Algorithms  \\ \hline
$\surd$ & $\surd$&$\times$ & $\times$ & $\times$ &$\times$  &  $\times$ & $\times$ \\\hline
$\surd$ & $\surd$&$\surd$ &  $\times$&  $\surd$ & $\times$ &    $\times$  &H1SXL (\ref{algo_sc}), H1SDRL (\ref{algo_H1SDRL})   \\\hline
$\surd$&$\surd$ & $\surd$  &$\surd$  &  $\surd$ & $\surd$ &   $\surd$  & H2SXL (\ref{algo_scx}), H2SDRL (\ref{algo_H2SDRL}) \\
 \hline
\end{tabular}
\end{center}
\end{table}

\section{Methodology}
 
In this section, we present our proposed methods.  We first study the identification of causal estimands under the model assumption of the SC method, which paves our path to the proposed heterogeneous one-side synthetic learners in Section \ref{sec:Estimator_one_side}. 
% under additional model assumptions
In Section \ref{sec:SCX}, when additional structures exist in the data, we can further extend our estimation procedure to construct heterogeneous two-side synthetic learners  that utilize more data and are expected to be more efficient. 
%, which achieves a faster convergence rate than the one-side learner. 
% (as detailed in Section \ref{sec:theory}). 
 
% \todohc{The current paper seems to discuss causal identification of HTT in Section 3.1 but propose estimators for HTE afterwards. We need to focus on either HTE or HTT. Just choose one and stick to it.}

% \subsection{Causal Identification}\label{sec:Causal identification}

\subsection{The Heterogeneous One-side Synthetic Learners}\label{sec:Estimator_one_side}

The key challenge of measuring treatment effects %\textcolor{blue}{HTT or HTE} 
is that the counterfactual outcomes can not be observed. 
%, i.e., one unit is either treated or under control. % defined in \eqref{eqn:HTE}
% causal inference is that the counter-factual outcomes can not be observed, e.g., the terms in our estimand definition \ref{eqn:HTT}. 
% following causal inference literature
The first step is to relate  the causal estimands with a statistical estimand that can be estimated from the observed data. 
To this end, we start to discuss the identifiability of {HTT} under the standard framework of SC,  which requires fewer assumptions  compared with other methods such as DiD \citep{abadie2021using}, and hence allows flexible data-generating processes and time-variant dependent predictors. 
Specifically, the SC model \citep{abadie2003economic, abadie2010synthetic} assumes that for the potential outcomes under control, we have:  
\begin{equation}\label{eq:scmulti}
Y_{i,t}(0) = \sum_{j=m+1}^{N} w_{i,j} Y_{j,t}(0) +  \varepsilon_{i,t}^{0}, 
\end{equation} % unobserved transitory shocks
for $ i = 1, 2, \cdots, m$ and $t = 1, 2, \cdots, T$, where $ \varepsilon_{i,t}^{0}$ is a random noise with $\Mean \left( \varepsilon_{i,t}^{0}\right)=0$ , $\Var \left( \varepsilon_{i,t}^{0}\right) = s^2 < \infty$. 
%We also assume $\Mean[\varepsilon_{i,t} | D_{i,t}, \boldsymbol{x}_{i,t}] = 0$. 
%can either be iid or, more generally, a stationary and weakly dependent process. 
Though it may look strong at first glance,  this model assumption is flexible and approximately valid under various data-generating  processes  including linear models, latent factor models, and autoregressive models. 
It is also well grounded in many real applications \citep{sabia2012effects, cavallo2013catastrophic,kreif2016examination,bayat2020synthetic,abadie2021using}. 
The underlying main idea is that, as the time series trends are typically driven by a few common factors, a weighted average of control units often provides a good approximation for the counterfactual outcome of the treated unit as if it were under control.   
The weighted average of control units is hence called a `synthetic' unit. 
Under this model, with Assumptions \ref{ass1}, \ref{SUTVA}, and \ref{ass:ci} (a) hold,
% and the resulting equation \eqref{mid_step}, 
the HTT function is identifiable from the data given the weights $\{w_{i,j}\}$ as 
\begin{equation}\label{eqn:final_identification}
\begin{aligned}
&\delta(\boldsymbol{x})  
 = \mathbb{E}\Big[Y_{i,t} - Y_{i,t}(0)\mid \boldsymbol{X}_i=\boldsymbol{x},D_{i,t} = 1\Big]\\
&= 
\mathbb{E}\Big[Y_{i,t} -\sum_{j=m+1}^{N} w_{i,j} Y_{j, t}  \mid \boldsymbol{X}_i=\boldsymbol{x},D_{i,t} = 1\Big],
\end{aligned}
\end{equation}
where $t> T_0$ corresponds to the experiment periods. 

 Inspired by 
% \citet{doudchenko2016balancing} and   
 \citet{chernozhukov2021exact}, we estimate the weights using restricted least squares: 
\begin{equation} \label{eq:psc0}
\begin{aligned}
& \widehat{ \boldsymbol{w}}_{i} = \arg \min _{(\mu, w)} \sum_{t=1}^{T_0}\left(Y_{i,t}  -\sum_{j=m+1}^{N} w_{i,j} Y_{j, t} \right)^2  \\
&\text { s.t. } \|\boldsymbol{w}_{i}\|_1 \leq 1, \boldsymbol{w}_{i} = \left( w_{i,m+1}, w_{i,m+2}, \cdots, w_{i,N}\right)^{\prime}.
\end{aligned}
\end{equation}
We estimate $Y_{i,t}(0)$ via
\begin{equation} \label{eq:estimator0}
	\widehat{Y}_{i,t}(0) = \sum_{j=m+1}^{N}\widehat{w}_{i,j} Y_{j,t},
\end{equation}
where $i = 1,2,\cdots, m$, $T_0+1 \leq t \leq T$. 

\textbf{Heterogeneous One-side Synthetic X-Learner (H1SXL). } 
%\todo{HC: This part should be H1SXL?}
The above derivation motivates a three-step approach, which is concisely summarized in Algorithm \ref{algo_sc}. 
In Step 1, we estimate the counterfactual outcomes without treatment for the treated units. 
In Step 2, we calculate the ITE $\tilde{\Delta}_{i,t}^{1}=Y_{i,t}^{1}- \widehat{Y}^{1}_{i,t}(0)
$ for every treated unit. Intuitively, the soundness of this Step  is based on the model assumption in Equation \eqref{eq:scmulti}.
Then we estimate the HTT function  $\widehat{\delta}(\boldsymbol{x})$ using imputed ITE in step 3. 
% and  the soundness of Step 3 is based on the identifiability of HTT \eqref{eqn:final_identification}. 
\begin{remark}
For a given treated unit, Steps 1 and 2 together can be regarded as the standard SC that estimates the ITE of a single unit. 
Our main contribution lies in elegantly extending this line of research to HTE estimation and proposing an efficient solution.
\end{remark}

\begin{algorithm}[tb]
\caption{Heterogeneous One-side Synthetic X-Learner (H1SXL)}\label{algo_sc}
\begin{algorithmic}
 \State \textbf{Input:} Data $\{\boldsymbol{X}_i,  D_{i,t}, Y_{i,t}\}_{i=1,2, \cdots, N, t=1,2,\cdots,T}$ 
 \State \textbf{Step 1:}  Estimate 
 $\widehat{\boldsymbol{w}}_{i}$ for any $i \le m$ based on Equation \eqref{eq:psc0} and calculate the couterfactual outcome without treatment for the treated units $\widehat{Y}_{i,t}(0)$ by Equation \eqref{eq:estimator0}.
 \State \textbf{Step 2:}	Impute the estimated ITE $\Delta_{i,t}$ as $\tilde{\Delta}_{i,t}^{1}=Y_{i,t}^{1}- \widehat{Y}_{i,t}(0)$ for any $i \le m$ and $ t \geq T_{0}+1$. 
 \State  \textbf{Step 3:} 	Estimate the HTT function as   $\widehat{\delta}(\boldsymbol{x})$, by regressing  $\tilde{\Delta}_{i,t}^{1}$ on $\boldsymbol{X}_i$, for any $i \le m$  and $ t \geq T_{0}+1$. Any user-specified regression methods can be applied in this step. 
\end{algorithmic}
\end{algorithm}

\begin{algorithm}[!tb]
   \caption{Heterogeneous One-side Synthetic Doubly Robust Learner (H1SDRL)}\label{algo_H1SDRL}
\begin{algorithmic}
\State {\bfseries Input:} Data $\{\boldsymbol{X}_i,  D_{i,t}, Y_{i,t}\}_{i=1,2, \cdots, N, t=1,2,\cdots,T}$ 
\State \textbf{Step 1.  ITE Estimation:} Estimate   $\tilde{\Delta}_{i,t}^{1}$ using Step 1 and Step 2 of Algorithm \ref{algo_sc}.
\State \textbf{Step 2. Sample Splitting:}  Denote $\widehat{O}_{i,t}=\{\boldsymbol{X}_i,  D_{i,t}, Y_{i,t}, \tilde{\Delta}_{i,t}^{D_{i,t}}\}$ where  $\tilde{\Delta}_{i,t}^{0}$ is missing. Split the sample into two parts as $\{\widehat{O}_{i,t}\}_{i\in \mathcal{S}_1}$ and  $\{\widehat{O}_{i,t}\}_{i\in \mathcal{S}_2}$. 
\State \textbf{Step 3. Nuisance training:} Based on data $\{\widehat{O}_{i,t}\}_{i\in \mathcal{S}_1}$, we can estimate the propensity score as $\widehat{e}(\cdot)$,  and the HTT function for the treated units as $\widehat{\delta}(\cdot)$ by regressing  $\tilde{\Delta}_{i,T}^{1}$ on $\boldsymbol{X}_i$ for any  $i\in \mathcal{S}_1$ and $D_{i,T}=1$.
\State \textbf{Step 4. Pseudo-outcome regression:} we construct the pseudo-outcome as
\begin{eqnarray}
\widehat{\phi}(\widehat{O}_{i,t}) = {1  \over \widehat{e}(\boldsymbol{X}_i)} \left\{\tilde{\Delta}_{i,T}^{1} - \widehat{\delta}(\boldsymbol{X}_i)\right\} +\widehat{\delta}(\boldsymbol{X}_i),
\end{eqnarray}
and regress it on $\boldsymbol{X}_i$ for $i \in \mathcal{S}_2$, which leads to
\begin{eqnarray}
\widehat{\delta}_{DR}(\boldsymbol{x}) = \widehat{\mathbb{E}}_n\{\widehat{\phi}(\widehat{O}_{i,t})|\boldsymbol{X}_i = \boldsymbol{x}\}.
\end{eqnarray}
\State \textbf{Step 4. Cross-fitting:}
An optional cross-fitting step can be applied according to \citet{kennedy2020optimal} by switching the role of $\mathcal{S}_1$ and $\mathcal{S}_2$, and then average the resulting estimates. 
\end{algorithmic}
\end{algorithm}

 \textbf{Heterogeneous One-side Synthetic Doubly Robust Learner (H1SDRL). } 
In case of model mis-specification , we propose  Heterogeneous One-side Synthetic Doubly Robust Learner (H1SDRL) based on  the estimated ITE 
%$\tilde{\Delta}_{i,t}^{1}=Y_{i,t}^{1}- \widehat{Y}^{1}_{i,t}(0)$ 
%for  treated units  
and  the DR-learner proposed by \citet{kennedy2020optimal}. The proposed method is summarized in Algorithm \ref{algo_H1SDRL}.

\subsection{The Heterogeneous Two-side Synthetic Learners } \label{sec:SCX}

After looking at our derivations in the previous part, one might wonder if we can implement  the same process in the opposite direction: for any unit $i$ in the \textit{control} group, we first impute $Y_{i,t}(1)$ as $\widehat{Y}_{i,t}(1)$ (i.e., a synthetic \textit{intervention} unit)  and then regress  $\widehat{Y}_{i,t}(1) - Y_{i,t}(0)$ on $\boldsymbol{X}_i$. 
Specifically,  we need the counterpart for Model \ref{eq:scmulti}: there exist weights $\{ v_{j,i}\}$ such that 
\begin{equation}\label{eq:scweight1}
\begin{aligned}
%Y_{i,t}(0) &= \sum_{j=1}^{m} w_{i,j} Y_{j, t}(0) + \varepsilon_{i,t},\\
Y_{j,t}(1) &= \sum_{i=1}^{m} v_{j,i} Y_{i, t}(1) +  \varepsilon_{j,t}^{1},
\end{aligned}
\end{equation}
hold for $j = m+1, m+2, \cdots, N$ and $t = 1, 2, \cdots, T$. 
Here,  $ \varepsilon_{j,t}^{1}$ is the random noise with $\Mean \left(  \varepsilon_{j,t}^{1} \right)=0$ , $\Var \left(  \varepsilon_{j,t}^{1} \right) = \tilde{s}^2 < \infty$. 
We estimate $v_{j,i}$ using
\begin{equation} \label{eq:psc2}
\begin{aligned}
\widehat{ \boldsymbol{v}}_{j} &= \arg \min _{(\mu, w)} \sum_{t=1}^{T_0}\left(Y_{j,t}  -\sum_{i=1}^{m} v_{j,i} Y_{i, t} \right)^2  \text { s.t. } \|\boldsymbol{v}_{j}\|_1 \leq 1,
\end{aligned}
\end{equation}
where $ \boldsymbol{v}_{j} = \left( v_{j,1}, v_{j,2}, \cdots, v_{j,m}\right)^{\prime}$.  We estimate $Y_{j,t}(1)$ via
\begin{equation} \label{eq:estimator1}
	\widehat{Y}_{j,t}(1) = \sum_{i=1}^{m}\widehat{v}_{j,i} Y_{i,t},
\end{equation}
where  $j = m+1, m+2, \cdots, N$, $T_0+1 \leq t \leq T$. 

The obtained estimator can then either be used alone or aggregated with the Heterogeneous One -side Synthetic Learners to construct a potentially more efficient one. 
Unfortunately, such an idea is not always feasible: 
intuitively, we only observe $\{Y_{i,t}(0)\}$ before $T_0 $ and hence can not expect to learn the relationship among  $\{Y_{i,t}(1)\}$.
% , which provides the foundation of constructing the synthetic \textit{control} outcomes. 

However, when additional structures exist, it is feasible to consider the opposite direction and hence extend our proposal.  The first thing we need is the stronger ignorability in Assumption \ref{ass:ci} (b). With this  assumption,  $Y_{i,t}(1)$ is identifiable from the data.
%the  treatment effects identified from the control units are statistically the same as in the whole population. 
As a bonus, we have the HTT $\delta(\boldsymbol{x})$ equals to the HTE $\tau(\boldsymbol{x})$, as summarized in Table \ref{tab:ci}. 
Besides 
%the ignorability
,  we also need to require 
%With additional assumption listed in Theorem \ref{thm:wexist1}, we have 
$ v_{j,i} =  w_{j,i} $, i.e.,
\begin{equation} \label{eq:scweight2} 
\begin{aligned}
Y_{j,t}(1)  &= \sum_{i=1}^{m} w_{j,i} Y_{i, t}(1)  +  \varepsilon_{j,t}^{1}, \\
Y_{j,t}(0) &= \sum_{i=1}^{m} w_{j,i} Y_{i, t}(0)  + \varepsilon_{j,t}^{0} , \\
\end{aligned}
\end{equation}
hold simultaneously for the control units $j = m+1, m+2, \cdots, N$. The above set of equations may seem strong at first glance , as it requires that the same set of linear relationships hold for both the potential outcomes for the control units under the treatment and the control. However, it is indeed reasonable in many applications. Here we discuss two of them, the aggregated data setup and the tensor factor model.

\textbf{Aggregated data. } 
One common case in real applications of SC is where the unit-level values are  aggregated  over individuals in  the unit, i.e., the so-called \textit{aggregated data}. 
For example, each unit can be a region where many individuals live  \citep{abadie2010synthetic, dube2015pooling, bayat2020synthetic}. % , as studied in
In this case, as argued in \citet{shi2021assumptions}, as long as 
%the individuals are homogeneous conditioned on their features, 
invariance assumptions across units and time were satisfied, 
under mild conditions, the linear factor model form holds for the unit-level outcomes.
%in both the treatment and control. 
See Appendix \ref{ap:modelass} for formal arguments.

\textbf{Tensor factor model. } 
Another common scenario is when the outcomes are driven by several common factors, i.e., when the tensor $\{Y_{i,t}(a)\}_{i \le N, t \le T, a \in \{0, 1\} }$ has a low-rank structure. 
Such a structure is well exploited in the tensor estimation literature and recently extended to casual panel data by \citet{Agarwal2021synthetic}. 
Specifically, the model states  that $Y_{i,t}(a) = \sum_{k=1}^K u_{t,k} v_{i,k} \xi_{a, k} + \varepsilon_{i,t}$, where $u_{t,k}$, $v_{i,k}$ and  $\xi_{a, k}$ are the latent time/unit/intervention factors, respectively. 
For example, \citet{bayat2020synthetic} applied this model to study the effect of different interventions in containing the COVID spread, across regions and over time. 

Given Equation \eqref{eq:scweight2},  we could use a similar procedure as in Section \ref{sec:Estimator_one_side} to estimate  the counterfactual treatment outcomes for the control units. Then with the estimated counterfactual outcomes for both treated units and control units, we are able to estimate the Individual Treatment Effect for each unit. We summarized the approach to estimating the Individual Treatment Effect, i.e., ITE in Algorithm \ref{algo_ite}.

\begin{algorithm}[tb]
   \caption{Identify the Individual Treatment Effect}\label{algo_ite}
\begin{algorithmic}
\State {\bfseries Input:} Data $\{\boldsymbol{X}_i,  D_{i,t}, Y_{i,t}\}_{i=1,2, \cdots, N, t=1,2,\cdots,T}$
% \State {\bfseries Input:} data $x_i$, size $m$
  \State \textbf{Step 1:}  Using Equation \eqref{eq:psc2} to estimate 
	$w_{i,j}$ as $\widehat{w}_{i,j}$ for any $i \le m$ and $j \ge m+1$,   and calculate the counterfactual outcome without treatment for the treated units by Equation \eqref{eq:estimator0}.
	\State \textbf{Step 2:} Estimate  $v_{j,i}$ as $\widehat{v}_{j,i}$ for any unit $i \le m$ and $j \ge m+1$ using Equation \eqref{eq:psc2},   and calculate the counterfactual outcomes $\widehat{Y}_{i,t}(1) $ under treatment for the control units  by Equation \eqref{eq:estimator1}.  
	\State \textbf{Step 3:}  Obtain the imputed individual treatment effects at each time point \begin{equation*}
		\begin{array}{l} \tilde{\Delta}_{i,t}^{1}:=Y_{i,t}-\widehat{Y}_{i,t}(0) \text{ for treated units},  \\ \tilde{\Delta}_{j,t}^{0}:=\widehat{Y}^{{\prime}}_{j,t}(1)-Y_{j,t} \text{ for control units}. \end{array}
	\end{equation*}
\end{algorithmic}
\end{algorithm}

After obtaining the ITE estimator from both sides,  we are able to construct our HTE estimator. In this section, we propose two Heterogeneous Two-side Synthetic Learners, namely Heterogeneous Two-side Synthetic X-Learner (H2SXL) and Heterogeneous Two-side Synthetic Doubly Robust Learner (H2SDRL).

\textbf{Heterogeneous Two-side Synthetic X-Learner (H2SXL). }
The basic idea of H2SXL can be described in three steps. 
In Step 1, we estimate ITE for each unit using Algorithm \ref{algo_ite}. Instead of employing two conditional mean functions of T-learners, the proposed H2SXL leverages counterfactual outcomes calculated using the SC Method, which enables us  to cope with temporal dependency in panel data.
In step 2, like all other Metalearners, we could use any off-the-shelf machine learning methods to learn the nuisance parameters, including the HTE functions for both treated units and control units, and the propensity score.
 In Step 3, we construct a weighted average with the HTE functions learned for both sides using  propensity weights. The weighting technique is useful when the percentages of the two groups are significantly unbalanced, similar to the original X-Leaner \citep{kunzel2019metalearners}. 
 We summarized the  proposed Heterogeneous Two-side Synthetic X-Learner (H2SXL)   in Algorithm \ref{algo_scx}. Such an estimator utilizes the opposite direction of the Heterogeneous One-side Synthetic X-Learner and is expected to gain more efficiency. 

\begin{algorithm}[tb]
   \caption{Heterogeneous Two-side Synthetic X-Learner (H2SXL)}\label{algo_scx}
\begin{algorithmic}
\State {\bfseries Input:} Data $\{\boldsymbol{X}_i,  D_{i,t}, Y_{i,t}\}_{i=1,2, \cdots, N, t=1,2,\cdots,T}$
  \State \textbf{Step 1. ITE Estimation:}  Estimate ITE $\tilde{\Delta}_{i,t}^{1}$ for the treated units and $\tilde{\Delta}_{j,t}^{0}$ for the control units using the method described in Algorithm \ref{algo_ite}.
	\State \textbf{Step 2. Nuisance training:}
	Estimate the propensity score for the treatment $\widehat{e}(\boldsymbol{x})$.
	And  estimate the HTE function for the treated units  as   
	\begin{equation*}
	\widehat{\tau}_{1}(\boldsymbol{x})= \widehat{\mathbb{E}}_n\{\tilde{\Delta}_{i,t}^{1}|\boldsymbol{X}_i = \boldsymbol{x}\}.
	\end{equation*}
%	$\widehat{\tau}_{0}(\boldsymbol{x})$, by regressing  $\tilde{\Delta}_{i,t}^{0}$ on $\boldsymbol{X}_i$,
	 for any  $i \leq  m$ and $t \geq T_{0}+1$; 

Similarly, estimate the HTE  for the control units  as   
	\begin{equation*}
	\widehat{\tau}_{0}(\boldsymbol{x})= \widehat{\mathbb{E}}_n\{\tilde{\Delta}_{j,t}^{0}|\boldsymbol{X}_i = \boldsymbol{x}\}.
	\end{equation*}
%	$\widehat{\tau}_{1}(\boldsymbol{x})$, by regressing  $\tilde{\Delta}_{j,t}^{1}$ on $\boldsymbol{X}_j$, 
	for $j \geq m+1$ and $t \geq T_{0}+1$.
	Any user-specified regression methods can be applied in this step;

  \State \textbf{Step 3. HTE Estimation: }  The estimated HTE function is $\widehat{\tau}(\boldsymbol{x})=\widehat{e}(\boldsymbol{x}) \widehat{\tau}_{0}(\boldsymbol{x})+(1-\widehat{e}(\boldsymbol{x})) \widehat{\tau}_{1}(\boldsymbol{x}) .$ 
\end{algorithmic}
\end{algorithm}

% The causal identification of the reversed direction requires a stronger ignorability as stated in Assumption \ref{ass:ci} (b). 

 \textbf{Heterogeneous Two-side Synthetic Doubly Robust Learner (H2SDRL). }
Similar as Section \ref{sec:Estimator_one_side}, in case of model mis-specification , we propose  Heterogeneous Two-side Synthetic Doubly Robust Learner (H2SDRL) in Algorithm \ref{algo_H2SDRL}.  In Step 3, the pseudo-outcome $\widehat{\phi}(\widehat{O}_{i,t}) $ is not a standard AIPW estimator and is doubly-robust 
for the conditional mean $\Mean \left[ \tilde{\Delta}_{i,T}\mid \boldsymbol{X}_i = \boldsymbol{x}_i\right]$.
%. The augmentation term 
%$(D_{i,T}-1/2){D_{i,T} - \widehat{e}(\boldsymbol{X}_i)\over {\widehat{e}(\boldsymbol{X}_i)\{1-\widehat{e}(\boldsymbol{X}_i)\}}} \left\{\tilde{\Delta}_{i,T}^{D_{i,T}} - \widehat{\tau}(\boldsymbol{X}_i)\right\}$ 
%ensures that it is a consistent estimator 
%as long as one of  $\widehat{e}(\cdot)$ and $\widehat{\tau}(\cdot)$ is consistently estimated.
 This allows us to estimate $\tau$ in a faster rate even if one of the nuisance estimates converge at a slower rate. 
The theoretical property is provided in Section \ref{sec:theory}.
%\textcolor{red}{Here, given $\tilde{\Delta}_{i,T}^{D_{i,T}}$ is consistent and the second-stage regression estimator $\widehat{\mathbb{E}}_n$ is stable (in terms of distance metric used), we have the rate doubly robustness of the proposed new estimator $\widehat{\tau}_{DR}$, i.e., $\widehat{\tau}_{DR}$ deviates from the truth by at most a product of errors in estimating the propensity score $e(\cdot)$ and the HTE function  \citep{kennedy2020optimal}. }

\begin{algorithm}[tb]
   \caption{Heterogeneous Two-side Synthetic Doubly Robust Learner (H2SDRL)}\label{algo_H2SDRL}
\begin{algorithmic}
\State {\bfseries Input:} Data $\{\boldsymbol{X}_i,  D_{i,t}, Y_{i,t}\}_{i=1,2, \cdots, N, t=1,2,\cdots,T}$ 
  \State \textbf{Step 1. ITE Estimation:}  Estimate ITE $\tilde{\Delta}_{i,t}^{1}$ for the treated units and $\tilde{\Delta}_{j,t}^{0}$ for the control units using the method described in Algorithm \ref{algo_ite}.
\State \textbf{Step 2. Sample Splitting:} 
Denote $\widehat{O}_{i,t}=\{\boldsymbol{X}_i, D_{i,T}, \tilde{\Delta}_{i,T}^{D_{i,T}}\}$. 
We split the sample into two parts as $\{\widehat{O}_{i,t}\}_{i\in \mathcal{S}_1}$ and  $\{\widehat{O}_{i,t}\}_{i\in \mathcal{S}_2}$. 
\State \textbf{Step 3. Nuisance training:} Based on data $\{\widehat{O}_{i,t}\}_{i\in \mathcal{S}_1}$, we can estimate the propensity score as $\widehat{e}(\cdot)$, and the HTE function $\widehat{\tau}(\cdot)$ by regressing  $\tilde{\Delta}_{i,T}$ on $\boldsymbol{X}_i$ for any  $i\in \mathcal{S}_1$.
\State \textbf{Step 4. Pseudo-outcome regression:} we construct the pseudo-outcome as 
\begin{eqnarray}
\widehat{\phi}(\widehat{O}_{i,t}) = (D_{i,T}-1/2){D_{i,T} - \widehat{e}(\boldsymbol{X}_i)\over {\widehat{e}(\boldsymbol{X}_i)\{1-\widehat{e}(\boldsymbol{X}_i)\}}} \left\{\tilde{\Delta}_{i,T}^{D_{i,T}} - \widehat{\tau}(\boldsymbol{X}_i)\right\} +\widehat{\tau}(\boldsymbol{X}_i),
\end{eqnarray}
and regress it on $\boldsymbol{X}_i$ for $i \in \mathcal{S}_2$, which leads to
\begin{eqnarray}
\widehat{\tau}_{DR}(\boldsymbol{x}) = \widehat{\mathbb{E}}_n\{\widehat{\phi}(\widehat{O}_{i,t})|\boldsymbol{X}_i = \boldsymbol{x}\}.
\end{eqnarray}
\State \textbf{Step 3. Cross-fitting:}
An optional cross-fitting step can be applied according to \citet{kennedy2020optimal} by switching the role of $\mathcal{S}_1$ and $\mathcal{S}_2$, and then average the resulting estimates. 
\end{algorithmic}
\end{algorithm}

\section{Theory}\label{sec:theory}

We provide theoretical guarantees for the proposed Heterogeneous Two-side Synthetic Learners in this section. Specifically, we first establish the convergence rate of the synthetic control outcomes, which equals the convergence rate of the estimated ITE. Then we  derive the convergence rate of the proposed Heterogeneous Two-side Synthetic Learners based on previous results of ITE estimation. The convergence rate of H1SXL and H1SDRL can be shown following the similar logic of proving the rate of H2SXL and H2SDRL so we omit that part for brevity. All the proofs can be found in the appendix.

%\textcolor{red}{The proofs assume that $T_0 > m \vee n$, however, our proposed learners work as well when there are fewer pre-pretreatment time points $T_0$ as shown in Section \ref{sec:sim}.} \todo{HC: shall we somewhere high-level mention that }

%We first state several technical assumptions as in SC.

\begin{assumption}[Positivity] \label{ass:Positivity}
There exists a positive constant $c$ such that $c<e(\boldsymbol{x})<1-c$ for any $\boldsymbol{x} \in \mathcal{X}$.
\end{assumption}

 With the  positivity assumption and mild technical assumptions listed in the appendix, we are able to prove the convergence rate for the synthetic control outcomes in  Equation \eqref{eq:estimator0} and Equation \eqref{eq:estimator1}, as stated in the following lemma.
 
%\begin{lemma}{(Convergence rate for the synthetic control outcomes)} \label{lm:sc}
%  Consider $\hat{\boldsymbol{w}}_i$ estimated using Equation \eqref{eq:estimator0} and $\hat{\boldsymbol{v}}_j$ estimated using Equation \eqref{eq:estimator1}  s.t. $\boldsymbol{w}_i , \boldsymbol{v}_j\in \mathcal{W}$, where $\mathcal{W}$ is a subset of $\left\{v:\|v\|_1 \leq K\right\}$ and $K$ is bounded. 
%  Assume Assumption \ref{ass:Positivity} and additional technical assumptions listed in the appendix about the moments and $\beta$-mixing  conditions of   $\varepsilon_{i,t} Y_{j,t} $ and $\tilde{\varepsilon}_{j,t} Y_{i,t}$ for $1\leq i\leq m$ and $m+1 \leq j \leq N$, the estimator enjoys the performance bounds: \\
%  \resizebox{\linewidth}{!}{$\frac{1}{T_1} \sum\limits_{t=T_0+1}^T\left[\hat{Y}_{i,t}(0)- \Mean \left \{Y_{i,t}(0)\right\} \right]^2=O_p([\log (T_1 \vee n)]^{(1+\tau) /(2 \tau)} T_1^{-1 / 2})$} \\
% for any $1 \leq i \leq m$, and \\
%   \resizebox{\linewidth}{!}{$\frac{1}{T_1} \sum_{t=T_0+1}^T\left[\hat{Y}_{j,t}(1)- \Mean \left\{Y_{j,t}(1)\right\} \right]^2=O_p([\log (T_1  \vee m)]^{(1+\tau) /(2 \tau)} T_1^{-1 / 2})$} \\
%  for any $m+1 \leq j\leq N$. 
%%  In the case of $n>T_1$, the estimator enjoys the performance bounds: $\frac{1}{T_1} \sum_{t=T_0+1}^T\left[\hat{Y}_{i,t}(0)- \Mean \left (Y_{i,t}(0)\right) \right]^2 =o_p\left(T_1^{-\alpha/2}\right).$
%\end{lemma}
\begin{lemma}[Convergence rate for the synthetic control outcomes]\label{lm:sc}
  Consider $\hat{\boldsymbol{w}}_i$ estimated using Equation \eqref{eq:estimator0} and $\hat{\boldsymbol{v}}_j$ estimated using Equation \eqref{eq:estimator1}. Assume $\boldsymbol{w}_i , \boldsymbol{v}_j\in \mathcal{W}$, where $\mathcal{W}$ is a subset of $\left\{v:\|v\|_1 \leq K\right\}$ and $K$ is bounded. 
  Assume Assumption \ref{ass:Positivity} and additional technical assumptions listed in the appendix about weakly dependent stationary properties of $Y_{i,t} $, $Y_{j,t} $ ,  $ \varepsilon_{i,t}^{0} Y_{j,t} $ and $ \varepsilon_{j,t}^{1} Y_{i,t}$ for $1\leq i\leq m$ and $m+1 \leq j \leq N$ hold, then the synthetic control  estimators enjoys the performance bounds: 
  \begin{equation*}
e_{i,t} \triangleq \Mean\{Y_{i,t}(0) \} - \widehat{Y}_{i,t}(0) = O_p(T_0^{-1/2}),1\leq i \leq  m,
\end{equation*}
and
\begin{equation*}
e_{j,t} \triangleq \Mean\{Y_{j,t}(1) \} - \widehat{Y}_{j,t}(1)   = O_p(T_0^{-1/2}), m+1 \leq j \leq N.
\end{equation*}
\end{lemma}

It is clear from Lemma \ref{lm:sc} that the convergence rate of one-side synthetic estimator of  potential outcomes is determined by the length of the pre-experiment period $T_0$ for fitting one-side SC models. 
%Similarly, the convergence rate of one-sided synthetic estimator of  potential outcome $Y_{j,t}(1)$ is determined by the number of treated units $m$ and the length of the post-experiment period $T_1$ for fitting one-side SC. 
Based on Lemma \ref{lm:sc}, we can quantify the estimation errors of ITE as stated in Algorithm \ref{algo_ite}. Built upon this, we are able to show the  convergence rate for Heterogeneous Two-side Synthetic X-Learner (H2SXL) in Theorem \ref{thm:H2SXL}.
% with following assumptions.

\begin{assumption} \label{ass:taures} 
%(a) Assume $\Delta_{i,t} = \delta(\boldsymbol{x}_{i}) + \overline{\epsilon}_{i,t}$ for unit $i$ and time $t$ with $D_{i,t} = 1$, where $ \overline{\epsilon}_{i,t}$ are i.i.d  given $\boldsymbol{X}_{i}=\boldsymbol{x}$ and $D_{i,t} = 1$, with $\mathbb{E}\left[\overline{\epsilon}_{i,t} \mid \boldsymbol{X}_{i}=\boldsymbol{x}, D_{i,t} = 1 \right]=0$ and $\operatorname{Var}\left[\overline{\epsilon}n_{i,t} \mid \boldsymbol{X}_{i}=\boldsymbol{x}, D_{i,t} = 1\right] \leq \sigma^{2}<\infty$. (b) 
Assume $\Delta_{i,t} = \tau(\boldsymbol{x}_{i}) + \overline{\epsilon}_{i,t}$, where $ \overline{\epsilon}_{i,t}$ are independent given $\boldsymbol{X}_{i}=\boldsymbol{x}$, with $\mathbb{E}\left[\overline{\epsilon}_{i,t} \mid \boldsymbol{X}_{i}=\boldsymbol{x}\right]=0$ and $\operatorname{Var}\left[\overline{\epsilon}_{i,t} \mid \boldsymbol{X}_{i}=\boldsymbol{x}\right] \leq \sigma^{2}<\infty$.
\end{assumption}

Assumptions \ref{ass:taures} is a standard  assumption to drive the convergence rate in the regression literature \citep[see e.g.,][]{lounici2008sup}. Specifically, 
Assumption \ref{ass:taures} requires  a conditional mean zero and bounded noise terms given the features and treatment information for all units.

\begin{theorem}[Convergence rate for H2SXL under linear assumption]\label{thm:H2SXL}Assume the treatment effect is linear, $\tau(\boldsymbol{x})=\boldsymbol{x}' \beta$.
% , with $\beta \in \mathbb{R}^{d}$ and $\overline{\epsilon}$ i.i.d  satisfying $\Mean(\overline{\epsilon}) = 0$ and  $\Mean(\overline{\epsilon}^2) = \sigma^2$  independent of $\varepsilon$. 
Assume  the assumptions for  Lemma \ref{lm:sc},   and Assumption \ref{ass:taures}, Assumption \ref{ass:xmatrix} hold with $|\widehat{e}(\boldsymbol{x}) - e(\boldsymbol{x})| = o_{p}(1)$, for any $\boldsymbol{x}$, the HTE estimator obtained by the proposed Heterogeneous Two-side Synthetic X-Learner (H2SXL) in  Algorithm \ref{algo_scx} has the following convergence rate:
\begin{equation*}
\mathbb{E} \left[ \left\{\hat{\tau}(\boldsymbol{x})- \tau(\boldsymbol{x}) \right\}^2 \mid \boldsymbol{x}  \right]  =   O_p\left(\left( n^{-1} + m^{-1}  + T_0^{-1}  \right)  T_1^{-1}   \right).
\end{equation*}
\end{theorem}

%\begin{theorem}{(Convergence rate for H2SXL})\label{thm2} Assume the treatment effect is linear, $\tau(\boldsymbol{x})=\boldsymbol{x}' \beta$.
%% , with $\beta \in \mathbb{R}^{d}$ and $\overline{\epsilon}$ i.i.d  satisfying $\Mean(\overline{\epsilon}) = 0$ and  $\Mean(\overline{\epsilon}^2) = \sigma^2$  independent of $\varepsilon$. 
%Assume  the assumptions for  Lemma \ref{lm:sc},   and Assumption \ref{ass:taures}, Assumption \ref{ass:xmatrix} hold with $|\widehat{e}(\boldsymbol{x}) - e(\boldsymbol{x})| = o_{p}(1)$, for any $\boldsymbol{x}$, the HTE estimator obtained by the proposed Heterogeneous Two-side Synthetic X-Learner (H2SXL) in  Algorithm \ref{algo_scx} has the following convergence rate:
%$ O_p\left(\frac{ [\log (T_1 \vee n)]^{(1+\tau) /(2 \tau)} +[\log (T_0 \vee m)]^{(1+\tau) /(2 \tau)} }{\sqrt{T_1}}+ \right. $  $\left.\frac{1}{m} +\frac{1}{n}  \right)$.
%\end{theorem}
%Note that, informally speaking, $e(\boldsymbol{x})/m$ and $(1-e(\boldsymbol{x}))/n$ are typically at the same order of $N$. 
%Therefore, compared with Theorem \ref{thm1}, when $m = o(T_0)$ and $m = o(N)$ (i.e., when the number of treated units is small, H2SL enjoys a faster convergence rate compared with H1SXL. 
%This clearly shows the benefit from additional utilizing the information in the control units. 

We make a few remarks for Theorem \ref{thm:H2SXL}. Clearly, the first two terms $n^{-1}T_1^{-1}$, $m^{-1}T_1^{-1}  $  is the estimation error due to the Step 2 of H2SXL, i.e., fitting the HTE function. The last   term $T_0^{-1} T_1^{-1}$is due to the estimation error from the SC method as characterized in Lemma \ref{lm:sc}. 
%In the case that the number of more  units is larger than time points, i.e. $m,n>T_1$,  the HTE function $\delta(\boldsymbol{x})$ can be consistently estimated by H2SXL if $T_0$ goes to infinity. 
% Using the similarly arguments in Lemma \ref{lm:sc}, we can show the following lemma that establish the estimation error of  the SC method on the opposite direction on treatment outcomes.

\begin{theorem}[Convergence Rate for H2SDRL] \label{thm:H2SDRL}
Assume the assumptions for  Lemma \ref{lm:sc},   and Assumption \ref{ass:taures} hold with the propensity score $\widehat{e}(\cdot)$  consistently estimated  in Step 1. Denote  $\tilde{\tau}(\boldsymbol{X}_i) = \Mean\left( \tilde{\Delta}_{i,T}^{D_{i,T}} \mid \boldsymbol{X}_i \right)$, and  assume $\widehat{\mathbb{E}}_n$ in Step 2 is a linear smoother of the form $\widehat{\mathbb{E}}_n\{\widehat{\phi}(\widehat{O}_{i,t}) \mid \boldsymbol{X}=\boldsymbol{x}\}=\sum_{i,t} h_{i,t}\left(\boldsymbol{x}\right) \widehat{\phi}(\widehat{O}_{i,t})$, then the HTE estimator obtained by the proposed Heterogeneous Two-side Synthetic Doubly Robust Learner (H2SDRL) in  Algorithm \ref{algo_H2SDRL} has the following convergence rate:
\begin{equation*}
\widehat{\tau}_{DR}(\boldsymbol{x})-\widehat{\mathbb{E}}\{\phi(\widehat{O}_{i,t}) \mid \boldsymbol{X}_i = \boldsymbol{x}\}
=\widehat{\mathbb{E}}_n\{\widehat{b}(\boldsymbol{X}_i) \mid \boldsymbol{X}_i=\boldsymbol{x}\}+o_{\mathbb{P}}\left(R_n^*(\boldsymbol{x})\right),
\end{equation*}
with 
\begin{equation*}
R_n^*(\boldsymbol{x})^2=\mathbb{E}\left[\widehat{\mathbb{E}}\{\phi(\widehat{O}_{i,t}) \mid \boldsymbol{X}_i = \boldsymbol{x}\} - \mathbb{E}\{\phi(\widehat{O}_{i,t}) \mid \boldsymbol{X}_i = \boldsymbol{x}\} \right]^2,
\end{equation*} 
\begin{equation*}
\widehat{b}(\boldsymbol{x})  = \frac{1}{2}   \left\{     \frac{1}{1- \widehat{e}(\boldsymbol{x})} -  \frac{1}{ \widehat{e}(\boldsymbol{x})}\right\}   \left\{ \widehat{e}(\boldsymbol{x})-e(\boldsymbol{x}) \right\}\left\{ \tilde{\tau}(\boldsymbol{x})   - \widehat{\tau}(\boldsymbol{x}) \right\}.
\end{equation*}
\end{theorem}
As stated in \citet{kennedy2020optimal},   linear smoothers have been widely studied in the literature, including linear regressions, local linear regression \citep{fan1993local}, and  random forests \citep{biau2012analysis}. In Theorem \ref{thm:H2SDRL}, we relax the Theorem 2 in \citet{kennedy2020optimal} by providing sufficient conditions.  In the special case of linear regressions, we give a corollary stated as following. 
   
%\todo{HC: need to high-level comment on Corollary 4.7. The difference between this DR rate with Theorem 4.5 lies in the MSE v.s. Bias? And thus $T_0$ is not shown in Corollary 4.7.}
\begin{coro} \label{coro:dr}
Suppose the assumptions of Theorem \ref{thm:H2SDRL} and Assumption \ref{ass:xmatrix}  hold. Further assume the treatment effect is linear, $\tau(\boldsymbol{x})=\boldsymbol{x}' \beta$ , $\widehat{\mathbb{E}}_n$  is a linear regression, and  $ |\widehat{e}(\cdot)-e(\cdot)| |\widehat{\tau}(\cdot)-\tilde{\tau}(\cdot)|= O_{\mathbb{P}} \left( (m+n)^{-1/2} T_1^{-1/2} \right)$. Then we have
\begin{equation*}
\mathbb{E} \left[ \left\{\widehat{\tau}_{DR}(\mathbf{x}) - \tau(\mathbf{x})\right\}^2 \mid \boldsymbol{x}  \right]  =   O_{\mathbb{P}} \left( \left\{(m+n)^{-1}   +T_0 ^{-1}\right\}T_1^{-1}\right).
\end{equation*}
\end{coro}
 
\section{Simulation\label{sec:sim}}
In this section, we conduct simulation studies to demonstrate the superior finite-sample performance of the proposed methods.  
Specifically, we consider the case of $N=50$ units with pre-treatment time $T_0=60$,  and post-treatment time $T_1=10$.  Each unit is treated with probability $\operatorname{Pr}(W_i=1)=0.5$. 
%The detailed data-generating process for potential outcomes under control $Y_{i,t}(0)$ can be found in the Appendix  \ref{app:simu}.

Within a unit $i$, generate a sample of $N_i \sim Unif(30,50)$ individuals. The support  $\mathbf{S}$ satisfies $\mathbf{S} \subset \{0,1,2,\cdots, 20\}$ with $|\mathbf{S}|=5$. 
The individual characteristics $S_{i,j}^{1},S_{i,j}^{2}$ are randomly sampled from $\mathbf{S}$ with replacement. The potential outcome under control is generated by $Y_{i,j,t}(0) =  b_{i,j} * f_{i,j,t} + e_{i,j,t},$ where $e_{i,j,t} \sim N(0,0.1^2)$ and $b_{i,j} \sim N(1+(S_{i,j}^{1}+S_{i,j}^{2})/40,1)$ is a constant for individual $j$ in unit $i$, and $t=1,2,\cdots, T_0+T_1$. The latent factor $f_{i,j,t}$ satisfies $f_{i,j,t} = 1.2 + 0.8* f_{i,j,t-1}+  u_{i,j,t}$ where   $f_{i,j,1} \sim N(2,0.1^2)$ and $u_{i,j,t}\sim N(0,0.1^2)$, $t = 2,3,\cdots, T_0+T_1 $. Then given $S_{i,j}^{1}$ and $S_{i,j}^{2}$, $Y_{i,j,t}(0)$ is only related to $f_{i,j,t}$ which have the same distribution across unit $i$. 
%Thus Assumption \ref{ICM0} is satisfied.

At the unit level, the potential outcome of $i$ is generated by $Y_{i,j}(0)  = \frac{1}{N_i} \sum_{j=1}^{N_i} Y_{i,j,t}(0) $. And the features of  unit $i$ are generated by $X_{i} =  \left( 1/N_i \sum_{j=1}^{N_i} \mathbf{I}\{S_{i,j}^{1}>12\}, 1/N_i \sum_{j=1}^{N_i} \mathbf{I}\{S_{i,j}^{1}>8\}\right).$

We consider three kinds of HTE functions: (1) linear HTE function with $\tau(x) = 0.6X_1+0.4X_2$; (2) non-linear HTE function with $\tau(x) = 0.6 X_1^2+0.4cos(X_2)$; (3) no heterogeneous  treatment effect with $\tau(x) = 0$. The potential outcomes of treated units under treatment are generated by $Y_{i,t}(1) = Y_{i,t}(0) + \tau(x_i)$.

We compare our proposed Heterogeneous Synthetic Learners with the following five baselines: (1) R-DiD proposed by \citet{nie2019nonparametric} which is the only existing HTE estimator for panel data to the best of our knowledge (implemented using  R-package \href{https://github.com/xnie/diffindiff}{diffindiff});  (2) X-Learner proposed by \citet{kunzel2019metalearners} which can be regarded as a base HTE estimator of our proposal but ignores the non-stationarity and the temporal dependency; and three other HTE estimators including the (3) R-Learner \citep{nie2021quasi}, (4) T-Learner \citep{athey2016recursive}, and (5) S-learner \citep{imai2013estimating} (implemented using  R-package  \href{https://github.com/xnie/rlearner}{rlearner}).  We evaluate these  methods by calculating the mean square error as MSE defined as $\sum_{i} \left(\widehat{\tau} (\boldsymbol{x}_i) -  \tau (\boldsymbol{x}_i)\right)^2$ for each run.

The experiment results for linear HTE functions are illustrated in 
Table \ref{tab:simu1_T060}, and additional simulation results for the other two settings can be found in Tables \ref{tab:simu2_T060} and \ref{tab:simu3_T060}. 
%The computing infrastructure used is a virtual machine at Google Colab with the Pro tier for the majority of the computations. 
%We notice that about 0.8\% out of the 1000 independent runs for RDiD fall with MSE greater than 10. Thus we only present successful runs of RDiD with MSE smaller than 10. 
The tables show that, in all simulation settings, the proposed methods significantly outperform RDiD and benchmarks that ignore temporal dependency, such as X-Learner, R-Learner, T-Learner, and S-Learner.

%\begin{table}[tb]
%\centering
%\caption{The mean and standard error (SE) of the calculated MSEs for HTE estimation under linear HTE functions. A total of  $m=50$ treated units,  with pre-treatment time $T_0=20$ and post-treatment time $T_1=10$.}
%\label{tab:simu1}
%\begin{tabular}{llllllllll}
%\toprule
% & H1SXL  & H2SXL & H1SDRL \\ 
%  \hline
%Mean & 0.0051 & 0.0040 & 0.0052 \\ 
%  SE & 0.0002 & 0.0002 & 0.0002 \\  
%    \midrule
%    \midrule 
%  & H2SDRL & RDiD & X-Learner \\ 
%  \hline
%Mean & 0.0041 & 0.9620 & 2.0166 \\ 
%  SE & 0.0002 & 0.3893 & 0.0071 \\
%   \midrule
%    \midrule
%   & R-Learner & T-Learner & S-Learner \\ 
%  \hline
%Mean & 1.3799 & 1.3802 & 1.3802 \\ 
%  SE & 0.0137 & 0.0138 & 0.0137 \\  
%      \bottomrule
%\end{tabular}
%\end{table}

%
\begin{table}[htb]
\centering
\caption{The mean and standard error (SE) of the calculated MSEs for HTE estimation under linear HTE functions. A total of  $m=50$ treated units,  with pre-treatment time $T_0=60$ and post-treatment time $T_1=10$.}
\label{tab:simu1_T060}
\begin{tabular}{llllllllll}
\toprule
 & H1SXL  & H2SXL & H1SDRL    & H2SDRL & RDiD\\ 
  \hline
Mean & 0.0037 & 0.0031 & 0.0039  & 0.0032 & 0.1562  \\ 
  SE & 0.0002 & 0.0001 & 0.0002 & 0.0001 & 0.0905  \\   
    \midrule
    \midrule
  & X-Learner   & R-Learner & T-Learner & S-Learner \\ 
  \hline
Mean & 0.2634  & 0.2366 & 0.2367 & 0.2367 \\ 
  SE & 0.0016 & 0.0048 & 0.0048 & 0.0048 \\   
      \bottomrule
\end{tabular}
\end{table}

\begin{table}[ht]
\centering
\caption{The mean and standard error (SE) of the calculated MSEs for HTE estimation under nonlinear HTE functions. A total of $m=50$ treated units,  with pre-treatment time $T_0=60$ and post-treatment time $T_1=10$.}
\label{tab:simu2_T060}
\begin{tabular}{llllllllll}
\toprule
 & H1SXL  & H2SXL & H1SDRL & H2SDRL & RDiD \\ 
  \hline
Mean & 0.0032 & 0.0030 & 0.0034 & 0.0029 & 0.1686 \\ 
  SE & 0.0001 & 0.0001 & 0.0002 & 0.0001 & 0.1002 \\  
    \midrule
    \midrule
   & X-Learner & R-Learner & T-Learner & S-Learner \\ 
  \hline
Mean & 0.2534 & 0.2355 & 0.2360 & 0.2358 \\ 
  SE & 0.0015 & 0.0048 & 0.0048 & 0.0048 \\ 
      \bottomrule
\end{tabular}
\end{table}

%\begin{table}[ht]
%\centering
%\caption{The mean and standard error (SE) of the calculated MSEs for HTE estimation under zero HTE functions. A total of $m=50$ treated units,  with pre-treatment time $T_0=20$ and post-treatment time $T_1=10$.}
%\label{tab:simu3}
%\begin{tabular}{llllllllll}
%\toprule
% & H1SXL  & H2SXL & H1SDRL & H2SDRL & RDiD\\ 
%  \hline
%Mean & 0.0043 & 0.0035 & 0.0046 & 0.0031 & 0.9122\\ 
%  SE & 0.0002 & 0.0002 & 0.0002 & 0.0001 & 0.4156\\ 
%    \midrule
%    \midrule
%   & X-Learner  & R-Learner & T-Learner & S-Learner \\ 
%  \hline
%Mean & 1.9849  & 1.3806 & 1.3805 & 1.3813 \\ 
%  SE & 0.0069  & 0.0137 & 0.0138 & 0.0138 \\ 
%      \bottomrule
%\end{tabular}
%\end{table}

\begin{table}[ht]
\centering
\caption{The mean and standard error (SE) of the calculated MSEs for HTE estimation under zero HTE functions. A total of $m=50$ treated units,  with pre-treatment time $T_0=60$ and post-treatment time $T_1=10$.}
\label{tab:simu3_T060}
\begin{tabular}{llllllllll}
\toprule
 & H1SXL  & H2SXL & H1SDRL & H2SDRL & RDiD\\ 
  \hline
Mean & 0.0031 & 0.0027 & 0.0033 & 0.0025 & 0.0866 \\ 
  SE & 0.0001 & 0.0001 & 0.0001 & 0.0001 & 0.0435 \\ 
    \midrule
    \midrule
   & X-Learner  & R-Learner & T-Learner & S-Learner \\ 
  \hline
MMean & 0.2511 & 0.2363 & 0.2367 & 0.2367 \\ 
  SE & 0.0015 & 0.0048 & 0.0048 & 0.0048 \\ 
      \bottomrule
\end{tabular}
\end{table}

\section{Real Data Application}

Following \citet{athey2021matrix}, we illustrate the advantages of our proposed method in real problems using a financial data set - 41266 minutes of data
ranging from April to August 2017 on 500 stocks which are available at \url{https://github.com/sebastianheinz/stockprediction/tree/master/01_data}. 
And we use their corresponding market value ($X_1$) and earnings per share ($X_2$) from Compustat as unit feature $\boldsymbol{X}$. We removed stocks with missing data, and there are a total of 459 stocks in our dataset.  
We randomly chose 100 units as treated units and the remaining 359 units as control units. And we randomly choose continuous 120 time points as pre-treatment time points, following 20 time points as post-treatment time points, and apply $\tau(\boldsymbol{X}) = 0.6/5000*X_1+0.4*X_2$ on the treated units after treatment assignment. 
We repeat the experiment 100 times and calculate the MSE for each run. We compare the performance of our proposed method with baselines including R-learner, T-learner, S-Learner, and X-Learner. The calculated MSEs for HTE estimation are summarized in Table \ref{tab:realdata}. It is shown that our proposed methods outperform benchmarks significantly in terms of MSE.

% \begin{table}[tb]
%     \centering
%     \caption{The mean and standard error (SE) of the calculated MSEs  for 100 independent runs of real data application. A total of $m=100$ treated units, $n=359$ control units with pre-treatment time $T_0=50$ and post-treatment time $T_1=20$.}
%     \label{tab:realdata}
% \begin{tabular}{llllllllll}
% \toprule
%  & H1SXL  & H2SXL & H1SDRL \\ 
%   \hline
% Mean & 1.2147 & 1.3360 & 1.3188 \\ 
%   SE & 0.0357 & 0.0304 & 0.0964 \\
%     \midrule
%     \midrule 
%  & H2SDRL & RDiD & X-Learner \\ 
%   \hline
% Mean & 2.9357 & 12.8205 & 5.6704 \\ 
%   SE & 0.0268 & 2.1614 & 0.1059 \\ 
%       \midrule
%     \midrule 
%   & R-Learner & T-Learner & S-Learner \\ 
%   \hline
% Mean & 28.5445 & 31.5549 & 30.7585 \\ 
%   SE & 1.4803 & 1.4281 & 1.3997 \\ 
%       \bottomrule   
% \end{tabular}
% \end{table}

\begin{table}[hb]
    \centering
    \caption{The mean and standard error (SE) of the calculated MSEs  for 100 independent runs of real data application. A total of $m=100$ treated units, $n=359$ control units with pre-treatment time $T_0=120$ and post-treatment time $T_1=20$.}
    \label{tab:realdata}
\begin{tabular}{llllllllll}
\toprule
 & H1SXL  & H2SXL & H1SDRL& H2SDRL & RDiD   \\ 
  \hline
Mean & 1.2407 & 1.3545 & 1.3495  & 2.9465 & 17.5936   \\ 
  SE & 0.0391 & 0.0308 & 0.1012 & 0.0296 & 3.0582\\ 
    \midrule
    \midrule 
  & X-Learner & R-Learner & T-Learner & S-Learner \\ \hline
Mean  & 3.1917 & 27.7543 & 30.6231 & 29.4438 \\ 
  SE  & 0.0482  & 1.4560 & 1.3977 & 1.3591 \\
      \bottomrule   
\end{tabular}
\end{table}

\section{Discussions}\label{sec:con}
In this paper, we systematically study the identification and estimation of heterogeneous treatment effects in panel data. 
When the ignorability assumption for $Y_{i,t}(0)$ holds,   we propose Heterogeneous One-side Synthetic X-Learner (H1SXL) and Heterogeneous One-side Synthetic Doubly Robust Learner (H1SDRL) for HTT  estimation. 
When a stronger ignorability assumption holds and appropriate data generation models apply,  we further extend the proposal to Heterogeneous Two-side Synthetic X-Learner (H2SXL) and Heterogeneous Two-side Synthetic Doubly Robust Learner (H2SDRL). 
There are several future directions to address potential limitations. 
%\textcolor{red}{Firstly, since  we  make the assumption that $T_0 > m \vee n$ in Section \ref{sec:theory}, a further study could extend the theory to the setting when there are fewer pre-pretreatment time points $T_0$. }
First,  we assume a linear model in Section \ref{sec:theory} to derive the convergence rate.  
Given the demonstrated good performance in the non-linear scenario, it is of interest to extend our theory to non-linear HTE functions. 
Second, the identifiability of HTT and HTE depends on the unverifiable ignorability assumption. 
Though how to relax this assumption is still an open question in causal inference, we may consider partial identifications. 
Finally, extending our framework to incorporating other causal panel data estimator such as augmented SC \citep{ben2021augmented} and  matrix completion \citep{athey2021matrix} is also desired.
    
\newpage

\bibliographystyle{agsm}
\bibliography{mycite}
% \bibliographystyle{icml2022}

%%%%%%%%%%%%%%%%%%%%%%%%%%%%%%%%%%%%%%%%%%%%%%%%%%%%%%%%%%%%%%%%%%%%%%%%%%%%%%%
%%%%%%%%%%%%%%%%%%%%%%%%%%%%%%%%%%%%%%%%%%%%%%%%%%%%%%%%%%%%%%%%%%%%%%%%%%%%%%%
% APPENDIX
%%%%%%%%%%%%%%%%%%%%%%%%%%%%%%%%%%%%%%%%%%%%%%%%%%%%%%%%%%%%%%%%%%%%%%%%%%%%%%%
%%%%%%%%%%%%%%%%%%%%%%%%%%%%%%%%%%%%%%%%%%%%%%%%%%%%%%%%%%%%%%%%%%%%%%%%%%%%%%%
\newpage
\appendix
\onecolumn

\counterwithin{figure}{section}
\counterwithin{table}{section}
\counterwithin{equation}{section}
% \counterwithin{definition}{section}

\section{A Brief Review of the Synthetic Control Methods}

In this section, we give a brief overview of the Synthetic Control Methods.
% \citet{abadie2003economic} and \citet{abadie2010synthetic}.
% and then introduce the SC Method for multiple treated units.

\textbf{The Classic Synthetic Control Method}
The original SC method proposed by \citet{abadie2010synthetic} considers an observational study, where there are finite $N = n+1$ units and $T = T_0 +1$ time periods with the first unit being treated.
% \todorw{This section should focus on 1. what is the parallel-trend assumption, 2. why SC is superior, and 3. the standard estimation procedure.}
It assumes that 
% the data follows parallel trends and 
the counterfactual outcome of the treated unit under no treatment $Y_{1,T_0+1}(0)$ can be well approximated by a weighted average of control units. The original SC method postulates the following regression model:
\begin{equation}\label{eq:scsingle}
Y_{1,t}(0) = \sum_{j=2}^{N} w_{j} Y_{j,t} + \epsilon_{1,t}, t = 1,2,\cdots,T_0+1,
% =\delta_{t}+\theta_{t} \sum_{j=2}^{N} w_{j} X_{j}+\lambda_{t} \sum_{j=2}^{N} w_{j} \mu_{j}+\sum_{j=2}^{N} w_{j} \overline{\epsilon}_{j, T_0+1}.
\end{equation} 
where the weights $\mathbf{w} = (w_2, w_3, \cdots,w_{N})^T$ are restricted to be nonnegative and sum to one to avoid extrapolation. $\epsilon_{1,t}$ are  error terms with zero mean, finite variance, and  uncorrelated with $Y_{j,t}$, i.e. $E\left(\epsilon_{1,t} Y_{j,t}\right)=0$ for $2 \leq j \leq N$.
% \todorw{I have some doubts on this is the model assumed by SC.}

Specifically, the weights are chosen to minimize  the difference  between the weighted average and the treated unit during the pre-treatment time period. We estimate the weights via the following constrained minimization problem: 
\begin{equation*}
\widehat{\mathbf{w}}=
%\arg \min _{\sum_{i=2}^{N}w_i = 1, W_i \geq 0} \sum_{t=1}^{T_{0}}\left[y_{1 t}- \sum_{i=2}^{N}w_i y_{i,t}\right]^{2} = 
\arg \min _{\sum_{i=2}^{N}w_i = 1, w_i \geq 0}  \sum_{t=1}^{T_0}\left(Y_{1,t}  -\sum_{j=2}^{N} w_{j} Y_{j, t} \right)^2,
% \left\|  \boldsymbol{Z}^{1}-  \sum_{i=1}^{N-1}w_i\boldsymbol{Z}^{0}  \right\|_2 ,
\end{equation*}
and then estimate $Y_{1,T_0+1}(0)$ via
$
	\widehat{Y}_{1,T_0+1}(0)= \sum_{i=2}^{N}\widehat{w}_i y_{i, T_0+1}.
$
This estimator is unbiased under latent factor models and  vector autoregressive models  with  certain conditions and the statistical inference can be  provided based on the idea of placebo studies \citep{abadie2010synthetic}.

\textbf{Relaxation of the Non-negative and Sum-to-One Assumptions} The non-negative and sum-to-one assumptions for the classic Synthetic Control method might be restrictive in real-life applications. A number of studies have shown that this problem can be overcome by adjusting the restrictions on the weights and using regression-based methods. \citet{ferman2019synthetic} suggested a demeaned Synthetic Control estimator incorporating an intercept into the SC problem while keeping the sum-to-one assumption. \citet{doudchenko2016balancing} further dropped all the restrictions and used an elastic net penalty.
%the coefficients sum to one restriction and developed a Modified Synthetic Control method that keeps the non-negative constraints.
  \citet{chernozhukov2021exact} replaced the original restrictions with a new one requiring the coefficients in a subset of an $l_1$ ball with a bounded radius and proved the consistency of the proposed estimator. Taking the imbalance of units' use as treatment and control units into account, \citet{bottmer2021design} added an additional set of restrictions on the weights to de-bias under randomized settings and provided a Modified Unbiased Synthetic Control estimator that allows intercepts and keeps all units used as controls as often as treated in expectation. From a different perspective, \citet{ben2021augmented} proposed an Augmented Synthetic Control Method to reduce bias due to imperfect pretreatment fit by adding a de-biasing term analogous to the standard doubly robust estimation \citep{robins1994estimation}.

\textbf{Synthetic Control Methods for Multiple Treated Units} With multiple treated units, practical challenges of non-unique solutions for the weights may arise and there are many alternative methods available for solving this problem. \citet{dube2015pooling} converted Synthetic Control estimates to elasticities and then aggregated these elasticities across events in a setting with recurring treatment and variable treatment intensity. \citet{robbins2017framework} proposed a framework for high-dimensional, micro-level data with multiple treated units and multiple outcome measures by calibrating the weights. \citet{xu2017generalized} generalized the original Method in \citet{abadie2003economic,abadie2010synthetic} to allow for several treated units by averaging effects across multiple units. \citet{abadie2021penalized} introduced an augmented Synthetic Control estimator with a penalty term to trade-off pairwise matching discrepancies with respect to the characteristics of each unit in the donor pool against matching discrepancies with respect to the characteristics of the units in the donor pool as a whole. The augmented Synthetic Control estimator utilizes control units with characteristics close to the treated  units in the space of matching variables and is able to reduce interpolation biases.

\section{SC with Aggregated Data} \label{ap:modelass}
% Model Assumptions in  Section \ref{sec:SCX}

In this section, we explain why the model we proposed for HTE estimation  in Section \ref{sec:SCX} is reasonable when the data is at the aggregated level. We firstly introduce the fine-grained potential outcomes framework for SC proposed by \citet{shi2021assumptions}.
At the individual level, let $Y_{i,j,t}(1)$ and $Y_{i,j,t}(0)$ be the potential outcome of interest for individual $j$ in unit $i$ on time period $t$ with and without treatment, respectively. 
% , where $i = 1,2, \cdots, n$, $j = 1,2, \cdots, l_i$ and $t = 1,2, \cdots, T_0+1$. 
Denote the vector of covariates of individual $j$ in unit $i$ on time period $t$ as  $\boldsymbol{Z}_{i,j,t}$. Follow \citet{shi2021assumptions}, we reason about individual-level variables $Y_{i,j,t}$ and $\boldsymbol{Z}_{i,j,t}$, but we never observe them,  as we only make observations at the unit level. We need the following two assumptions for the latent factor model used in SC.

\begin{assumption}(Independent Causal Mechanism) \label{ICM0}
Conditional on the covariate $\boldsymbol{Z}$, the potential outcome $Y(0)$ is independent of the unit distribution $i$. For unit distribution $i$ at time $t$, the joint distribution of $Y(0)$  and $\boldsymbol{Z}$ satisfies
\begin{equation*}
\begin{aligned}
P_{i,t}(\boldsymbol{Z}, Y(0)) & =P_{i,t}(\boldsymbol{Z}) P_{t}(Y(0) \mid \boldsymbol{Z}),\\
% P_{i,t}(\boldsymbol{Z}, Y(1)) & =P_{i,t}(\boldsymbol{Z}) P_{t}(Y(1) \mid \boldsymbol{Z}).
\end{aligned}
\end{equation*}
\end{assumption}
\begin{assumption}(Stable Distributions) \label{Stable}
Decompose the covariates into two subsets $\boldsymbol{Z}=\{\boldsymbol{U}, \boldsymbol{S}\}$. Let $S$ denote the subset that differentiates the target from the selected donors, i.e., its distribution in the target group is different from its distribution in the selected donor groups. We assume that, for all groups, the distribution of $S$ does not change for all time periods $t $,
\begin{equation*}
P_{i,t}(\boldsymbol{Z})=P_{i}(\boldsymbol{S}) P_{t}(\boldsymbol{U} \mid \boldsymbol{S}) .
\end{equation*}
\end{assumption}

Assumption \ref{ICM0} requires that conditional on $\boldsymbol{Z}$,  there are  no unmeasured variables that affect both the unit affiliation and potential outcomes.  Assumption \ref{Stable} says that the covariates $\boldsymbol{Z}$ can be divided into two sets, one is  "within-individual" variables which are the same among different units but change over time, and the other is  "among-individual" variables $\boldsymbol{S}$ which do not change over time but are different among units.

% Thus, it is natural to assume the following linear  factor model for the expected potential outcome $Y_{i,t}(0)$, which is commonly adopted in the literature of Synthetic Control \citep{abadie2003economic,abadie2010synthetic,abadie2021using,li2020statistical},
%  \begin{equation} \label{scmodel0}
% Y_{i,t}(0) =\delta_{t}+\theta_{t} \boldsymbol{X}_{i}+\lambda_{t} \mu_{i}+\overline{\epsilon}_{i,t},
% \end{equation}
% where $\delta_{t}$ is a time fixed effect,  $\theta_{t}$ is a  time-varying coefficient vector,$ \mu_{i}$ is a vector of time-invariant unobserved predictor variables with time-varying coefficients $\lambda_{t}$,  and $\overline{\epsilon}_{i,t}$ are unobserved transitory shocks with zero mean.

With two additional assumptions, we are able to proof the following Theorem \ref{thm:wexist}.

\begin{assumption}(Sufficiently Similar Donors)  \label{SimilarDonors}
Let $D$ be the set of donors used to construct the synthetic control and $S$ be the minimal invariant set for the target and the selected donors. The donors are sufficiently similar if the cardinality of the donor set is greater than or equal to the cardinality of the minimal invariant set,
\begin{equation*}
|D| \geq|S| \text {. }
\end{equation*}
\end{assumption}
\begin{assumption}(Target Donors Overlap)\label{TargetDonorsOverlap}
Let $\left\{s_{1}, \ldots, s_{R}\right\}$ be the support of $S$. There exists at least one donor distribution $i$ in the selected donor set $D$, where $P_{i}(S=s)>0$.	
\end{assumption}

\begin{theorem}{(Extension of Theorem 1 in \citet{shi2021assumptions} )} \label{thm:wexist}
Under Assumption \ref{ICM0}, \ref{Stable}, \ref{SimilarDonors} and \ref{TargetDonorsOverlap}, for any $i=1,2,\cdots,m$, there exist weights $\boldsymbol{w}_{i} = (w_{i,m+1}, w_{i,m+2}, \cdots,w_{i,N})$  such that 
\begin{equation}\label{eq:scweight0}
Y_{i,t}(0) = \sum_{j=m+1}^{N} w_{i,j} Y_{j, t}(0),  P_{i}( \boldsymbol{S}=\boldsymbol{s})= \sum_{j=m+1}^{N} w_{i,j} P_{j}( \boldsymbol{S}=\boldsymbol{s}), \forall s \in S,
\end{equation}
hold simultaneously.
\end{theorem}

\textbf{Proof}:  Under Assumption \ref{ICM0}  and Assumption \ref{Stable} , for any unit $i\in \{1,2,\cdots, N\}$, we can rewrite the expected potential outcome,
\begin{equation*}
\begin{aligned}
 \mathbb{E}\left[Y_{i,t}(0)\right] &=\sum_{\boldsymbol{z}}\mathbb{E}[Y(0)_{i,t} \mid \boldsymbol{Z}=\boldsymbol{z}] P_{i,t}(\boldsymbol{Z}=\boldsymbol{z})\\ & \overset{(B.1)}{=}\sum_{\boldsymbol{z}}\mathbb{E}_{t}[Y(0) \mid \boldsymbol{Z}=\boldsymbol{z}] P_{i,t}(\boldsymbol{Z}=\boldsymbol{z}) \\
 & \overset{(B.2)}{=}\sum_{(\boldsymbol{s},\boldsymbol{u})}\mathbb{E}_{t}[Y(0) \mid \boldsymbol{S}=\boldsymbol{s}, \boldsymbol{U}=\boldsymbol{u} ]   P_{i}(\boldsymbol{S}=\boldsymbol{s}) P_{t}(\boldsymbol{U}=\boldsymbol{u} \mid \boldsymbol{S}=\boldsymbol{s})\\
 & = \sum_{\boldsymbol{s}} P_{i}(\boldsymbol{S}=\boldsymbol{s}) \sum_{\boldsymbol{u}}\mathbb{E}_{t}[Y(0) \mid \boldsymbol{S}=\boldsymbol{s}, \boldsymbol{U}=\boldsymbol{u} ]  P_{t}(\boldsymbol{U}=\boldsymbol{u} \mid \boldsymbol{S}=\boldsymbol{s}) \\
 &   =\sum_{\boldsymbol{s}} P_{i}( \boldsymbol{S}=\boldsymbol{s})\mathbb{E}_{t}[Y(0) \mid \boldsymbol{S}=\boldsymbol{s}] , 
\end{aligned}
\end{equation*}
i.e. 
\begin{equation} \label{eq:linearY0}
\mathbb{E}\left[Y_{i,t}(0)\right]  =\sum_{\boldsymbol{s}} \mathbb{E}_{t}[Y(0) \mid \boldsymbol{S}=\boldsymbol{s}] P_{i}( \boldsymbol{S}=\boldsymbol{s}).
\end{equation}
With time $t$ fixed, we could regard the above expression as a system of linear equations. We observe both $\mathbb{E}\left[Y_{j,t}(0)\right]$ and  $P_{j}( \boldsymbol{S}=\boldsymbol{s})$ for  any $j=m+1, m+2, \cdots, N$ in the donor pool, with  $\mathbb{E}_{t}[Y(0) \mid \boldsymbol{S}=\boldsymbol{s}]$ unknown. By Assumption \ref{TargetDonorsOverlap}, for any $\boldsymbol{s}\in \boldsymbol{S}$, the unknown quantity  $\mathbb{E}_{t}[Y(0) \mid \boldsymbol{S}=\boldsymbol{s}]$ is in at least one of equations. Assumption \ref{SimilarDonors} says that the 
number of  independent equations $|\boldsymbol{D}|$ is no less than the number of unknowns $|\boldsymbol{S}|$. Therefore, the unknown quantities $\mathbb{E}_{t}[Y(0) \mid \boldsymbol{S}=\boldsymbol{s}],\boldsymbol{s}\in \boldsymbol{S}$ can be solved. Thus, for fixed $t$ and any $\boldsymbol{s}\in \boldsymbol{S}$, there exist weight  $\boldsymbol{v}_{t} = (v_{m+1,t}, v_{m+2,t}, \cdots,v_{N,t})$, such that 
\begin{equation*}
\mathbb{E}_{t}[Y(0) \mid \boldsymbol{S}=\boldsymbol{s}] = \sum_{j=m+1}^{N} v_{j,t}  \mathbb{E}\left[Y_{j,t}(0)\right],
\end{equation*}
where the weights $v_{j,t}$ are functions of $P_{j}( \boldsymbol{S}=\boldsymbol{s})$, $j=m+1, \cdots, N$. Since $P_{j}( \boldsymbol{S}=\boldsymbol{s})$ is invariant with time,  the weights $v_{j,t}$ are invariant with time too. Therefore, for any time $t$, 
 there exist weights $\boldsymbol{v} = (v_{m+1}, v_{m+2}, \cdots,v_{N})$  such that 
\begin{equation*}
\mathbb{E}_{t}[Y(0) \mid \boldsymbol{S}=\boldsymbol{s}] = \sum_{j=m+1}^{N} v_{j}  \mathbb{E}\left[Y_{j,t}(0)\right],
\end{equation*}
Thus, combined with equation \ref{eq:linearY0}, for any time $t$, and  any treated unit, i.e.  $i = 1,2, \cdots, m$, we have 
\begin{equation} 
\begin{aligned}
\mathbb{E}\left[Y_{i,t}(0)\right]  & =\sum_{\boldsymbol{s}} \mathbb{E}_{t}[Y(0) \mid \boldsymbol{S}=\boldsymbol{s}] P_{i}( \boldsymbol{S}=\boldsymbol{s}) =\sum_{\boldsymbol{s}}  \left\{ \sum_{j=m+1}^{N} v_{j}  \mathbb{E}\left[Y_{j,t}(0)\right] \right\}  P_{i}( \boldsymbol{S}=\boldsymbol{s})\\
& = \sum_{j=m+1}^{N} v_{j} \left\{ \sum_{\boldsymbol{s}} P_{i}( \boldsymbol{S}=\boldsymbol{s}) \right\} \mathbb{E}\left[Y_{j,t}(0)\right]   \triangleq \sum_{j=m+1}^{N} w_{i,j}  \mathbb{E}\left[Y_{j,t}(0)\right],
\end{aligned}
\end{equation}
where $w_{i,j}=v_{j} \left\{ \sum_{\boldsymbol{s}} P_{i}( \boldsymbol{S}=\boldsymbol{s}) \right\}$.

Further, combined with equation \ref{eq:linearY0}, we have
\begin{equation*}
\begin{aligned}
 \mathbb{E}\left[Y_{i,t}(0)\right]   &  = \sum_{j=m+1}^{N} w_{i,j}  \mathbb{E}\left[Y_{j,t}(0)\right] = \sum_{j=m+1}^{N} w_{i,j}  \sum_{\boldsymbol{s}} \mathbb{E}_{t}[Y(0) \mid \boldsymbol{S}=\boldsymbol{s}] P_{j}( \boldsymbol{S}=\boldsymbol{s}) \\
 & =  \sum_{\boldsymbol{s}} \mathbb{E}_{t}[Y(0) \mid \boldsymbol{S}=\boldsymbol{s}] \sum_{j=m+1}^{N} w_{i,j}  P_{j}( \boldsymbol{S}=\boldsymbol{s}),
\end{aligned}
\end{equation*}
which follows
\begin{equation*}
\sum_{\boldsymbol{s}} \mathbb{E}_{t}[Y(0) \mid \boldsymbol{S}=\boldsymbol{s}] P_{i}( \boldsymbol{S}=\boldsymbol{s}) = \mathbb{E}\left[Y_{i,t}(0)\right]  =  \sum_{\boldsymbol{s}} \mathbb{E}_{t}[Y(0) \mid \boldsymbol{S}=\boldsymbol{s}] \sum_{j=m+1}^{N} w_{i,j}  P_{j}( \boldsymbol{S}=\boldsymbol{s}),
\end{equation*}
i.e. 
\begin{equation*}
\sum_{\boldsymbol{s}} \mathbb{E}_{t}[Y(0) \mid \boldsymbol{S}=\boldsymbol{s}]  \left\{ P_{i}( \boldsymbol{S}=\boldsymbol{s})-\sum_{j=m+1}^{N} w_{i,j}  P_{j}( \boldsymbol{S}=\boldsymbol{s}) \right\} = 0.
\end{equation*}
Since the above equation holds for any form of $ \mathbb{E}_{t}[Y(0) \mid \boldsymbol{S}=\boldsymbol{s}] $ and any time $t$, we have 
\begin{equation*}
 P_{i}( \boldsymbol{S}=\boldsymbol{s})=\sum_{j=m+1}^{N} w_{i,j}  P_{j}( \boldsymbol{S}=\boldsymbol{s}).
\end{equation*}
The proof is completed.

Next, based on the above theorem, with an additional assumption for causal identification of the opposite direction we could further proof Theorem \ref{thm:wexist1}.
\begin{assumption}(Independent Causal Mechanism) \label{ICM1}
Conditional on the covariate $\boldsymbol{Z}$, the potential outcome  $Y(1)$ are independent of the unit distribution $i$. For unit distribution $i$ at time $t$, the joint distribution of $ Y(1)$ and $\boldsymbol{Z}$ satisfies
\begin{equation*}
\begin{aligned}
 P_{i,t}(\boldsymbol{Z}, Y(1)) & =P_{i,t}(\boldsymbol{Z}) P_{t}(Y(1) \mid \boldsymbol{Z}).
\end{aligned}
\end{equation*}
\end{assumption}

\begin{theorem} \label{thm:wexist1}
With Assumptions required by Theorem \ref{thm:wexist} hold with Assumption \ref{ICM1}, for any $j=m+1,m+2,\cdots,N$,  there exist weights $\boldsymbol{w}_{j} = (w_{j,1}, w_{j,2}, \cdots,w_{j,m})$  such that 
\begin{equation*}
\Mean \left[Y_{j,t}(1)\right]  = \sum_{i=1}^{m} w_{j,i} \Mean \left[Y_{i, t}(1)\right] ,\Mean \left[Y_{j,t}(0)\right]  = \sum_{i=1}^{m} w_{j,i} \Mean \left[Y_{i, t}(0)\right] ,  P_{j}( \boldsymbol{S}=\boldsymbol{s})= \sum_{i=1}^{m} w_{j,i} P_{i}( \boldsymbol{S}=\boldsymbol{s}), \forall s \in S,
\end{equation*}
hold simultaneously.
\end{theorem}

\textbf{Proof}:  The proof of Theorem \ref{thm:wexist1} is based on the proof of Theorem \ref{thm:wexist}.  First,  under the Assumptions required by Theorem \ref{thm:wexist}, there exist weights $\boldsymbol{w}_{j} = (w_{j,1}, w_{j,2}, \cdots,w_{j,m})$   such that 
\begin{equation*}
\Mean \left[Y_{j,t}(0)\right] = \sum_{i=1}^{m} w_{j,i} \Mean \left[Y_{i, t}(0)\right] ,  P_{j}( \boldsymbol{S}=\boldsymbol{s})= \sum_{i=1}^{m} w_{j,i} P_{i}( \boldsymbol{S}=\boldsymbol{s}), \forall s \in S,
\end{equation*}
hold simultaneously.  Next with Assumption \ref{ICM1} holds, we could similarly prove that there exist weights $\boldsymbol{w}'_{j} = (w'_{j,1}, w'_{j,2}, \cdots,w'_{j,m})$   such that 
\begin{equation*}
\Mean \left[Y_{j,t}(1)\right]  = \sum_{i=1}^{m} w'_{j,i} \Mean \left[Y_{i, t}(1)\right] ,  P_{j}( \boldsymbol{S}=\boldsymbol{s})= \sum_{i=1}^{m} w'_{j,i} P_{i}( \boldsymbol{S}=\boldsymbol{s}), \forall s \in S,
\end{equation*}
hold simultaneously. 

Recall that $w_{j,i}=v_{i} \left\{ \sum_{\boldsymbol{s}} P_{j}( \boldsymbol{S}=\boldsymbol{s}) \right\}$ with $\boldsymbol{v} = (v_{1}, v_{2}, \cdots,v_{m})$ solved by $\mathbb{E}_{t}[Y(0) \mid \boldsymbol{S}=\boldsymbol{s}] = \sum_{j=1}^{m} v_{i}  \mathbb{E}\left[Y_{i,t}(0)\right],$
using the linear equations similar as Equation \ref{eq:linearY0}, i.e. 
$\mathbb{E}\left[Y_{i,t}(0)\right]  =\sum_{\boldsymbol{s}} \mathbb{E}_{t}[Y(0) \mid \boldsymbol{S}=\boldsymbol{s}] P_{i}( \boldsymbol{S}=\boldsymbol{s}),$
and the weights $w_{j,i}$ are functions of only  $P_{i}( \boldsymbol{S}=\boldsymbol{s})$, $i=1, \cdots, m$. And similarly we have $w_{j,i}'=v_{i}' \left\{ \sum_{\boldsymbol{s}} P_{j}( \boldsymbol{S}=\boldsymbol{s}) \right\}$ with $\boldsymbol{v}' = (v'_{1}, v'_{2}, \cdots,v'_{m})$ solved by $\mathbb{E}_{t}[Y(1) \mid \boldsymbol{S}=\boldsymbol{s}] = \sum_{i=1}^{m} v'_{i}  \mathbb{E}\left[Y_{i,t}(1)\right],$
using 
$\mathbb{E}\left[Y_{i,t}(1)\right]  =\sum_{\boldsymbol{s}} \mathbb{E}_{t}[Y(1) \mid \boldsymbol{S}=\boldsymbol{s}] P_{i}( \boldsymbol{S}=\boldsymbol{s}),$
and the weights $w'_{j,i}$ are functions of only  $P_{i}( \boldsymbol{S}=\boldsymbol{s})$, $i=1, \cdots, m$. Since the structure of the two sets of linear equations are the same, we have $\boldsymbol{w}'_{j} =\boldsymbol{w}_{j}$. Then the conclusion follows immediately.

% Under Assumption \ref{ICM0},  Assumption \ref{Stable}  and Assumption \ref{ICM1}, similarly as the Equation  \ref{eq:linearY0},  we can rewrite the expected potential outcome under treatment as
% \begin{equation}  \label{eq:linearY1}
% \begin{aligned}
%  \mathbb{E}\left[Y_{i,t}(1)\right] &   =\sum_{\boldsymbol{z}}\mathbb{E}[Y(1)_{i,t} \mid \boldsymbol{Z}=\boldsymbol{z}] P_{i,t}(\boldsymbol{Z}=\boldsymbol{z}) \\
%  & =\sum_{\boldsymbol{z}}\mathbb{E}_{t}[Y(1) \mid \boldsymbol{Z}=\boldsymbol{z}] P_{i,t}(\boldsymbol{Z}=\boldsymbol{z}) =\sum_{\boldsymbol{s}} \underbrace{\mathbb{E}_{t}[Y(1) \mid \boldsymbol{S}=\boldsymbol{s}]}_{\lambda'_{t}} \underbrace{P_{i}( \boldsymbol{S}=\boldsymbol{s})}_{\gamma_{i}}.
% \end{aligned}
% \end{equation}
% Similarly,  it is natural to assume a linear  factor model for the expected potential outcome $Y_{i,t}(1)$:
%  \begin{equation} \label{scmodel1}
% Y_{i,t}(1) =\delta_{t}'+\theta_{t}' \boldsymbol{X}_{i}+\lambda_{t}' \mu_{i}'+\overline{\epsilon}_{i,t}'.
% \end{equation}
% Recall that $\Delta_{i,t} = Y_{i,t}(1) - Y_{i,t}(0)$, it is immediate that 
% \begin{equation*}
% \Delta_{i,t} = \delta_{t}'-\delta_{t}+ \left(\theta_{t}'-\theta_{t} \right) \boldsymbol{X}_{i}+ \left(\lambda_{t}' \mu_{i}'-\lambda_{t} \mu_{i}\right)+\overline{\epsilon}_{i,t}'-\overline{\epsilon}_{i,t},
% \end{equation*}
% which implies that $\Delta_{i,t}$ is linear in $ \boldsymbol{X}_{i}$. Thus the model we proposed in Section \ref{sec:SCX} is reasonable.

%\input{AlgorithmCompare}

\section{Proofs} \label{sec:supproofs}
In this section, we provide all the proofs of theoretical results in the paper 'Heterogeneous Synthetic Learner for Panel Data'. Additional technical assumptions are listed as following.

\begin{assumption} \label{ass:Li1}
(1) Denote 
%$\mathbf{Y}_t^{trt} = \left( Y_{1,t}, Y_{2,t}, \cdots, Y_{m,t}\right)$ as the potential outcome under control for all the treated units at time $t$ and similarly denote
 $\mathbf{Y}_t^{con} = \left( Y_{m+1,t}, Y_{m+2,t}, \cdots, Y_{m+n,t}\right)^{\prime}$. Assume $\left\{\mathbf{Y}_t^{con}\right\}_{t=1}^{T_0}$ is a weakly dependent stationary process so that laws of large number holds: $T_0^{-1} \sum_{t=1}^{T_0}\mathbf{Y}_t^{con} \stackrel{p}{\rightarrow} E\left(\mathbf{Y}_t^{con}\right)$ 
and $\left(\mathbf{H}^{con \prime} \mathbf{H}^{con} / T_0\right) \equiv T_0^{-1} \sum_{t=1}^{T_0} \mathbf{Y}_t^{con} \mathbf{Y}_t^{con \prime} \stackrel{p}{\rightarrow} E\left(\mathbf{Y}_t^{con} \mathbf{Y}_t^{con \prime}\right)$, where $ E\left(\mathbf{Y}_t^{con}\mathbf{Y}_t^{con\prime}\right)$ is positive definite, and $\mathbf{H}^{con}$ is the $T_0\times n$ matrix with its $t^{t h}$ row given by $\mathbf{Y}_t^{con \prime}$. 
%Let $\phi=\lim _{T_1, T_2 \rightarrow \infty} \sqrt{T_2 / T_1}$, then $\phi$ is a finite non-negative constant.
\\ (2) Denote 
 $\mathbf{Y}_t^{trt} = \left( Y_{1,t}, Y_{2,t}, \cdots, Y_{m,t}\right)^{\prime}$. Assume $\left\{\mathbf{Y}_t^{trt}\right\}_{t=1}^{T_0}$ is a weakly dependent stationary process so that laws of large number holds: $T_0^{-1} \sum_{t=1}^{T_0}\mathbf{Y}_t^{trt} \stackrel{p}{\rightarrow} E\left(\mathbf{Y}_t^{trt}\right)$ 
and $\left(\mathbf{H}^{trt \prime} \mathbf{H}^{trt} / T_0\right) \equiv T_0^{-1} \sum_{t=1}^{T_0} \mathbf{Y}_t^{trt} x\mathbf{Y}_t^{trt \prime} \stackrel{p}{\rightarrow} E\left(\mathbf{Y}_t^{trt} \mathbf{Y}_t^{trt \prime}\right), E\left(\mathbf{Y}_t^{trt}\mathbf{Y}_t^{trt\prime}\right)$ is positive definite, where $\mathbf{H}^{trt}$ is the $T_0\times m$ matrix with its $t^{t h}$ row given by $\mathbf{Y}_t^{trt \prime}$. 
\end{assumption}
\begin{assumption}\label{ass:Li2}
(1) For any $j=m+1,\cdots,N$, $\left\{\varepsilon_{j,t}^{1} \right\}_{t=1}^T$ is zero mean, serially uncorrelated and satisfies $T_0^{-1 / 2} \sum_{t=1}^{T_0} \mathbf{Y}_t^{con}\varepsilon_{j,t}^{1} \stackrel{d}{\rightarrow}$ $N\left(0, \Sigma_1\right)$, where $\Sigma_1=E\left( \left(\varepsilon_{j,t}^{1}\right)^2 \mathbf{Y}_t^{con}\mathbf{Y}_t^{con\prime}\right)$
\\(2)For any $i=1,\cdots, m$, $\left\{\varepsilon_{i,t}^{0} \right\}_{t=1}^T$ is zero mean, serially uncorrelated and satisfies $T_0^{-1 / 2} \sum_{t=1}^{T_0} \mathbf{Y}_t^{trt}\varepsilon_{i,t}^{0}  \stackrel{d}{\rightarrow}$ $N\left(0, \Sigma_2\right)$, where $\Sigma_2=E\left(\left( \varepsilon_{i,t}^{0} \right)^2 \mathbf{Y}_t^{trt}\mathbf{Y}_t^{trt\prime}\right)$
\end{assumption}

\begin{assumption}\label{ass:xmatrix}
The eigenvalues of the sample covariance matrixes for both treated units and control units are well conditioned, in specific, there exists some positive constants $\lambda_1, \lambda_2$ such that

(a) $ 0 < \lambda_1 < \lambda_{\min}\left( ({1}/{m}) \sum_{i=1}^{m} \boldsymbol{x}_{i}\boldsymbol{x}_{i}^{\prime} \right) <  \lambda_{\max}\left(({1}/{m}) \sum_{i=1}^{m}   \boldsymbol{x}_{i}\boldsymbol{x}_{i}^{\prime}  \right) < \lambda_2 $;

(b) $ 0 <  \lambda_1 < \lambda_{\min}\left( ({1}/{n}) \sum_{i=m+1}^{N} \boldsymbol{x}_{i}\boldsymbol{x}_{i}^{\prime} \right) <  \lambda_{\max}\left(({1}/{n}) \sum_{i=m+1}^{N}   \boldsymbol{x}_{i}\boldsymbol{x}_{i}^{\prime}  \right) < \lambda_2 $.

(c) $ 0 <  \lambda_1 < \lambda_{\min}\left( ({1}/{N}) \sum_{i=1}^{N} \boldsymbol{x}_{i}\boldsymbol{x}_{i}^{\prime} \right) <  \lambda_{\max}\left(({1}/{N}) \sum_{i=1}^{N}   \boldsymbol{x}_{i}\boldsymbol{x}_{i}^{\prime}  \right) < \lambda_2 $.
\end{assumption}

Assumptions \ref{ass:xmatrix} is a  standard  assumption to drive the convergence rate in the regression literature \citep[see e.g.,][]{lounici2008sup}. Specifically, 
Assumption \ref{ass:xmatrix} is a technical assumption required to bound the tail of regression estimator so a sharp rate can be achieved.

\subsection{Proof for Lemma \ref{lm:sc}}

Let $Z_1$ denote the limiting distribution of $\sqrt{T_0}\left(\widehat{\mathbf{w}}_{i}^{O L S}-\mathbf{w}_{i}\right)$ i.e., $\mathcal{N}\left(0, s^2 T_0^{-1}E\left(\mathbf{Y}_t^{con} \left\{ \mathbf{Y}_t^{con\prime}\right) \right\}^{-1} \right)$ for $i=1,2,\cdots,m$, then under the Assumptions \ref{ass:Li1} to \ref{ass:Li2} presented , by Theorem 3.2 in \cite{li2020statistical}, we have
\begin{equation*}
\sqrt{T_0}\left(\widehat{\mathbf{w}}_{i} -\mathbf{w}_{i} \right) \stackrel{d}{\rightarrow} \Pi_{T_{\Lambda, \mathbf{w}_{i}}} Z_1 ,
\end{equation*}
with 
\begin{equation} \label{eq:PiDef}
\Pi_{\Lambda} \theta=\arg \min _{\lambda \in \Lambda}(\theta-\lambda)^{\prime} E\left(\mathbf{Y}_t^{con}\mathbf{Y}_t^{con\prime}\right)(\theta-\lambda),
\end{equation}
and $T_{\Lambda, \mathbf{w}_{i}}$ being the 'tangent cone' of $\Lambda$ at $\mathbf{w}_{i}$ defined as
\begin{equation*}
T_{\Lambda, \mathbf{w}_{i}}=\overline{U_{\alpha \geq 0} \alpha\left\{\Lambda-\Pi_{\Lambda} \mathbf{w}_{i}\right\}}.
\end{equation*}
Similarly, let $Z_2$ denote the limiting distribution of $\sqrt{T_0}\left(\widehat{\mathbf{v}}_{j}^{O L S}-\mathbf{v}_{j}\right)$  $\mathcal{N}\left(0, s^2 T_0^{-1}E\left(\mathbf{Y}_t^{trt} \left\{ \mathbf{Y}_t^{trt\prime}\right) \right\}^{-1} \right)$ for $j=m+1, m+2 \cdots, m+n$, then under the Assumptions \ref{ass:Li1} to \ref{ass:Li2} presented ,  by Theorem 3.2 in \cite{li2020statistical}, we have
\begin{equation*}
\sqrt{T_0}\left(\widehat{\mathbf{v}}_{j} -\mathbf{v}_{j} \right) \stackrel{d}{\rightarrow} \Pi_{T_{\Lambda, \mathbf{v}_{j}}} Z_2 ,
\end{equation*}
Thus, for $i=1,2,\cdots,m$ and $t=T_0+1, T_0+2, \cdots, T$, we have
\begin{equation*}
e_{i,t} = \Mean\{Y_{i,t}(0) \} - \widehat{Y}_{i,t}(0)  = \mathbf{Y}_t^{con} \left\{ \mathbf{w}_{i} - \widehat{\mathbf{w}}_{i} \right\} = O_p(T_0^{-1/2}),
\end{equation*}
and for $j=m+1,m+2,\cdots,m+n$ and $t=T_0+1, T_0+2, \cdots, T$, we have
\begin{equation*}
e_{j,t} = \Mean\{Y_{j,t}(1) \} - \widehat{Y}_{j,t}(1)  = \mathbf{Y}_t^{trt} \left\{ \mathbf{v}_{j} - \widehat{\mathbf{v}}_{j} \right\} = O_p(T_0^{-1/2}).
\end{equation*}

\subsection{Proof for Theorem \ref{thm:H2SXL}}
\textbf{Firstly, we aim to proof the following lemma.}
\begin{lemma} \label{lemma:oneside}
Under the assumptions of Theorem \ref{thm:H2SXL}, we have 
\begin{equation*}
\mathbb{E} \left[ \left\{\hat{\tau}_1(\boldsymbol{x})- \tau(\boldsymbol{x}) \right\}^2 \mid \boldsymbol{x}  \right]   =   \mathcal{O}_p\left(  \left( m^{-1} + T_0^{-1}  \right) T_1^{-1}\right).
\end{equation*}
\end{lemma}

Note that
\begin{equation}  \label{eq:DeltaError}
\tilde{\Delta}_{i,t}^{1} =Y_{i,t}(1)- \widehat{Y}_{i,t}(0) = Y_{i,t}(1) -  Y_{i,t}(0) + Y_{i,t}(0)- \Mean\{Y_{i,t}(0) \}+ \Mean\{Y_{i,t}(0) \} - \widehat{Y}_{i,t}(0) = \Delta_{i,t}^{1} + \varepsilon_{i,t}^{0}  + e_{i,t},
\end{equation}
where $Y_{i,t}(1) -  Y_{i,t}(0)  = \Delta_{i,t}^{1} $, $Y_{i,t}(0)- \Mean\{Y_{i,t}(0) \} = \varepsilon_{i,t}^{0} $, and $\Mean\{Y_{i,t}(0) \} - \widehat{Y}_{i,t}(0) = e_{i,t}$.
By Equation \eqref{eq:DeltaError},
\begin{equation}  \label{eq:DeltaDecom}
\tilde{\Delta}_{i,t}^{1}=  \tau(\boldsymbol{x}_{i}) + \overline{\epsilon}_{i,t} +\varepsilon_{i,t}^{0} +e_{i,t}  =  \boldsymbol{x}_{i}^{\top} \beta + \overline{\epsilon}_{i,t}+\varepsilon_{i,t}^{0}  +e_{i,t}.
\end{equation}
where $\overline{\epsilon}_{i,t}$ is independent of $e_{i,t}$ since $ \overline{\epsilon}_{i,t}$ are i.i.d  given $\boldsymbol{X}_{i}=\boldsymbol{x}$ and $D_{i,t} = 1$ by Assumption \ref{ass:taures}.
 We estimate $\beta$ using an OLS estimator,
\begin{equation*}
\hat{\beta}=\left(\sum_{t=T_0+1}^{T}  \sum_{i=1}^{m} \boldsymbol{x}_{i}\boldsymbol{x}_{i}^{\prime} \right)^{-1} \sum_{t=T_0+1}^{T} \sum_{i=1}^{m} \boldsymbol{x}_{i} \tilde{\Delta}_{i,t}^{1}.
\end{equation*}
 We decompose $\hat{\tau}_0(\boldsymbol{x})- \tau(\boldsymbol{x}) $ into two independent error terms:
 \begin{equation*}
\begin{aligned}
 \hat{\tau}_1(\boldsymbol{x})- \tau(\boldsymbol{x})  & = \boldsymbol{x}^{\top} \hat{\beta}-\boldsymbol{x}^{\top} \beta = \boldsymbol{x}^{\top} (\beta-\hat{\beta}) =  \boldsymbol{x}^{\top}\left(\sum_{t=T_0+1}^{T}  \sum_{i=1}^{m} \boldsymbol{x}_{i}\boldsymbol{x}_{i}^{\prime} \right)^{-1} \sum_{t=T_0+1}^{T} \sum_{i=1}^{m} \boldsymbol{x}_{i} \left( \overline{\epsilon}_{i,t}+\varepsilon_{i,t}^{0} +e_{i,t}\right) \\
 &= \boldsymbol{x}^{\top} \left(\sum_{t=T_0+1}^{T}  \sum_{i=1}^{m} \boldsymbol{x}_{i}\boldsymbol{x}_{i}^{\prime} \right)^{-1} \sum_{t=T_0+1}^{T} \sum_{i=1}^{m} \boldsymbol{x}_{i}  \left(\overline{\epsilon}_{i,t}+\varepsilon_{i,t}^{0}\right) + \boldsymbol{x}^{\top}\left(\sum_{t=T_0+1}^{T}  \sum_{i=1}^{m} \boldsymbol{x}_{i}\boldsymbol{x}_{i}^{\prime} \right)^{-1} \sum_{t=T_0+1}^{T} \sum_{i=1}^{m} \boldsymbol{x}_{i} e_{i,t}.
\end{aligned}
 \end{equation*}
 Denote $\boldsymbol{X}^{1} = \left( \boldsymbol{X}_{1},\boldsymbol{X}_{1}, \cdots, \boldsymbol{X}_{m}\right)$, $ \boldsymbol{\overline{\epsilon}}_{t} = \left( \overline{\epsilon}_{1,t},  \overline{\epsilon}_{2,t}, \cdots,  \overline{\epsilon}_{m,t} \right)^{\prime}$, $ \boldsymbol{\varepsilon}_{t}^{0} = \left( \varepsilon_{1,t}^{0},  \varepsilon_{2,t}^{0}, \cdots,  \varepsilon_{m,t}^{0} \right)^{\prime}$ , $ \boldsymbol{e}_{t} = \left( e_{1,t},  e_{2,t}, \cdots,  e_{m,t} \right)^{\prime}$,  then 
\begin{equation*}
\hat{\tau}_1(\boldsymbol{x})- \tau(\boldsymbol{x}) =  \boldsymbol{x}^{\top} \left(\sum_{t=T_0+1}^{T} \boldsymbol{X}^{1}  \boldsymbol{X}^{1 \prime} \right)^{-1} \sum_{t=T_0+1}^{T}\boldsymbol{X}^{1} \left(\boldsymbol{\overline{\epsilon}}_{t}+ \boldsymbol{\varepsilon}_{t}^{0} \right)  + \boldsymbol{x}^{\top}\left(\sum_{t=T_0+1}^{T} \boldsymbol{X}^{1}  \boldsymbol{X}^{1 \prime} \right)^{-1} \sum_{t=T_0+1}^{T}\boldsymbol{X}^{1}   \boldsymbol{e}_{t}.
\end{equation*} 
Thus, by the triangle inequality, we have
\begin{equation} \label{eq:term}
\mathbb{E} \left[ \left\{\hat{\tau}_1(\boldsymbol{x})- \tau(\boldsymbol{x}) \right\}^2 \mid \boldsymbol{x}  \right]  \leq\|\boldsymbol{x}\|_2^2 \mathbb{E}\left[\left\| \left(\sum_{t=T_0+1}^{T} \boldsymbol{X}^{1}  \boldsymbol{X}^{1 \prime} \right)^{-1} \sum_{t=T_0+1}^{T}\boldsymbol{X}^{1}   \left(\boldsymbol{\overline{\epsilon}}_{t}+ \boldsymbol{\varepsilon}_{t}^{0} \right)  \right\|_2^2+\left\| \left(\sum_{t=T_0+1}^{T} \boldsymbol{X}^{1}  \boldsymbol{X}^{1 \prime} \right)^{-1} \sum_{t=T_0+1}^{T}\boldsymbol{X}^{1}   \boldsymbol{e}_{t} \right\|_2^2 \right].
\end{equation}
For the first term, we have
\begin{equation*}
\begin{aligned}
\mathbb{E} \left\| \left(\sum_{t=T_0+1}^{T} \boldsymbol{X}^{1}  \boldsymbol{X}^{1 \prime} \right)^{-1} \sum_{t=T_0+1}^{T}\boldsymbol{X}^{1}  \left(\boldsymbol{\overline{\epsilon}}_{t}+ \boldsymbol{\varepsilon}_{t}^{0} \right)  \right\|_2^2 & =\mathbb{E} \left\| \sum_{t=T_0+1}^{T} \left(T_1 \boldsymbol{X}^{1}  \boldsymbol{X}^{1 \prime} \right)^{-1} \boldsymbol{X} ^{1}  \left(\boldsymbol{\overline{\epsilon}}_{t}+ \boldsymbol{\varepsilon}_{t}^{0} \right) \right\|_2^2\\
& = \frac{1}{T_1^{2}}\mathbb{E}\left\| \sum_{t=T_0+1}^{T} \left(\boldsymbol{X}^{1}   \boldsymbol{X}^{1 \prime} \right)^{-1} \boldsymbol{X}^{1}  \left(\boldsymbol{\overline{\epsilon}}_{t}+ \boldsymbol{\varepsilon}_{t}^{0} \right) \right\|_2^2 . \\
\end{aligned}
\end{equation*}
Utilizing the properties of the trace, we can write
\begin{equation*}
\begin{aligned}
& \mathbb{E} \left\| \left(\sum_{t=T_0+1}^{T} \boldsymbol{X}^{1}  \boldsymbol{X}^{1 \prime} \right)^{-1} \sum_{t=T_0+1}^{T}\boldsymbol{X}^{1} \left(\boldsymbol{\overline{\epsilon}}_{t}+ \boldsymbol{\varepsilon}_{t}^{0} \right)   \right\|_2^2 \\
& =  \frac{1}{T_1^{2}}\sum_{t=T_0+1}^{T} \mathbb{E} \left[ tr \left\{\left(\boldsymbol{\overline{\epsilon}}_{t}+ \boldsymbol{\varepsilon}_{t}^{0} \right)^{\prime} \boldsymbol{X}^{1 \prime} \left(\boldsymbol{X}^{1}   \boldsymbol{X}^{1 \prime} \right)^{-1}  \left(\boldsymbol{X}^{1}   \boldsymbol{X}^{1 \prime} \right)^{-1} \boldsymbol{X}^{1} \left(\boldsymbol{\overline{\epsilon}}_{t}+ \boldsymbol{\varepsilon}_{t}^{0} \right)  \right\}\right] \\
& =  \frac{1}{T_1^{2}}\sum_{t=T_0+1}^{T} \mathbb{E} \left[ tr \left\{ \boldsymbol{X}^{1 \prime} \left(\boldsymbol{X}^{1}   \boldsymbol{X}^{1 \prime} \right)^{-1}  \left(\boldsymbol{X}^{1}   \boldsymbol{X}^{1 \prime} \right)^{-1} \boldsymbol{X}^{1} \left(\boldsymbol{\overline{\epsilon}}_{t}+ \boldsymbol{\varepsilon}_{t}^{0} \right) \left(\boldsymbol{\overline{\epsilon}}_{t}+ \boldsymbol{\varepsilon}_{t}^{0} \right)^{\prime} \right\}\right] \\
& =  \frac{1}{T_1^{2}}\sum_{t=T_0+1}^{T} \mathbb{E} \left[ tr \left\{ \boldsymbol{X}^{1 \prime} \left(\boldsymbol{X}^{1}   \boldsymbol{X}^{1 \prime} \right)^{-1}  \left(\boldsymbol{X}^{1}   \boldsymbol{X}^{1 \prime} \right)^{-1} \boldsymbol{X}^{1} \left(\boldsymbol{\overline{\epsilon}}_{t}+ \boldsymbol{\varepsilon}_{t}^{0} \right) \left(\boldsymbol{\overline{\epsilon}}_{t}+ \boldsymbol{\varepsilon}_{t}^{0} \right)^{\prime} \right\}\right]. \\
\end{aligned}
\end{equation*}	
Therefore, we have
\begin{equation*} 
\mathbb{E} \left\| \left(\sum_{t=T_0+1}^{T} \boldsymbol{X}^{1}  \boldsymbol{X}^{1 \prime} \right)^{-1} \sum_{t=T_0+1}^{T}\boldsymbol{X}^{1} \left(\boldsymbol{\overline{\epsilon}}_{t}+ \boldsymbol{\varepsilon}_{t}^{0} \right)  \right\|_2^2	=  \frac{1}{T_1^{2}}\sum_{t=T_0+1}^{T}  tr\left\{  \boldsymbol{X}^{1 \prime} \left(\boldsymbol{X}^{1}   \boldsymbol{X}^{1 \prime} \right)^{-1}  \left(\boldsymbol{X}^{1}   \boldsymbol{X}^{1 \prime} \right)^{-1} \boldsymbol{X}^{1}\mathbb{E} \left( \left(\boldsymbol{\overline{\epsilon}}_{t}+ \boldsymbol{\varepsilon}_{t}^{0} \right) \left(\boldsymbol{\overline{\epsilon}}_{t}+ \boldsymbol{\varepsilon}_{t}^{0} \right)^{\prime} \right) \right\}.
\end{equation*}
Since $\boldsymbol{\overline{\epsilon}}_{t}$ is independent of $\boldsymbol{\varepsilon}_{t}^{0}$, we have $\Mean \left(\overline{\epsilon}_{i,t}+ \varepsilon_{i,t}^{0}\right)^2=\Mean \left(\overline{\epsilon}_{i,t}^2\right)+\Mean \left\{ \left(\varepsilon_{i,t}^{0}\right)^2\right\}=\sigma^2 + s^2$, thus we have 
\begin{equation*}
\begin{aligned}
\mathbb{E} \left\{ \left(\boldsymbol{\overline{\epsilon}}_{t}+ \boldsymbol{\varepsilon}_{t}^{0} \right) \left(\boldsymbol{\overline{\epsilon}}_{t}+ \boldsymbol{\varepsilon}_{t}^{0} \right)^{\prime} \right\}  & = diag \left\{ \Mean \left(\overline{\epsilon}_{1,t}^2\right)+\Mean\left\{  \left( \varepsilon_{1,t}^{0}\right)^2 \right\}, \Mean \left(\overline{\epsilon}_{2,t}^2\right)+\Mean\left\{  \left( \varepsilon_{2,t}^{0}\right)^2 \right\},\cdots,\Mean \left(\overline{\epsilon}_{m,t}^2\right)+\Mean\left\{  \left( \varepsilon_{m,t}^{0}\right)^2 \right\} \right\} = \left( \sigma^2 + s^2\right) \boldsymbol{I}_{m} 
\end{aligned}
\end{equation*}
for any $i=1,2,\cdots, m$.
Therefore 
\begin{equation} \label{eq:trinequal}
\begin{aligned}
\mathbb{E} \left\| \left(\sum_{t=T_0+1}^{T} \boldsymbol{X}^{1}  \boldsymbol{X}^{1 \prime} \right)^{-1} \sum_{t=T_0+1}^{T}\boldsymbol{X}^{1} \left(\boldsymbol{\overline{\epsilon}}_{t}+ \boldsymbol{\varepsilon}_{t}^{0} \right)  \right\|_2^2 & = \frac{1}{T_1^{2}}\sum_{t=T_0+1}^{T}  tr\left\{  \boldsymbol{X}^{1 \prime} \left(\boldsymbol{X}^{1}   \boldsymbol{X}^{1 \prime} \right)^{-1}  \left(\boldsymbol{X}^{1}   \boldsymbol{X}^{1 \prime} \right)^{-1}\boldsymbol{X}^{1} \left( \sigma^2 + s^2\right) \boldsymbol{I}_{m} \right\} \\ 
&  =  \frac{\sigma^2+s^2}{T_1^{2}} \sum_{t=T_0+1}^{T}  tr\left\{  \boldsymbol{X}^{1 \prime} \left(\boldsymbol{X}^{1}   \boldsymbol{X}^{1 \prime} \right)^{-1}  \left(\boldsymbol{X}^{1}   \boldsymbol{X}^{1 \prime} \right)^{-1} \boldsymbol{X}^{1} \right\}. \\ 
\end{aligned}
\end{equation}
Note that by Assumption \ref{ass:xmatrix}
\begin{equation}\label{eq:trace}
\begin{aligned}
& tr \left\{ \boldsymbol{X}^{\prime} \left(\boldsymbol{X}^{1}   \boldsymbol{X}^{1 \prime} \right)^{-1}  \left(\boldsymbol{X}^{1}   \boldsymbol{X}^{1 \prime} \right)^{-1} \boldsymbol{X}^{1}    \right\} = tr \left\{ \boldsymbol{X}^{1}  \boldsymbol{X}^{1 \prime} \left(\boldsymbol{X}^{1}   \boldsymbol{X}^{1 \prime} \right)^{-1}  \left(\boldsymbol{X}^{1}   \boldsymbol{X}^{\prime} \right)^{-1}  \right\} = tr \left\{  \left(\boldsymbol{X}^{1}   \boldsymbol{X}^{1 \prime} \right)^{-1}  \right\}\\
&  = \sum_{i} \frac{1}{\lambda_{i}\left(\boldsymbol{X}^{1}   \boldsymbol{X}^{1 \prime} \right)} = \sum_{i} \frac{1}{ m \lambda_{i}\left( \frac{1}{m} \boldsymbol{X}^{1}   \boldsymbol{X}^{1 \prime} \right)} \leq \frac{d}{ m\lambda_{\min}\left( \frac{1}{m} \boldsymbol{X}^{1}   \boldsymbol{X}^{1 \prime} \right)} \leq \frac{d}{m c_1},
\end{aligned}
\end{equation}
thus we have
\begin{equation} \label{eq:term1}
\mathbb{E} \left\| \left(\sum_{t=T_0+1}^{T} \boldsymbol{X}^{1}  \boldsymbol{X}^{1 \prime} \right)^{-1} \sum_{t=T_0+1}^{T}\boldsymbol{X}^{1} \left(\boldsymbol{\overline{\epsilon}}_{t}+ \boldsymbol{\varepsilon}_{t}^{0} \right)  \right\|_2^2 \leq  \frac{\sigma^2+s^2}{T_1^{2}} \sum_{t=T_0+1}^{T} \frac{d}{m \lambda_1} =   \frac{d \left(\sigma^2 + s^2 \right)}{m \lambda_1 T_1}.
\end{equation}
For the second term, by the triangle inequality, we have
\begin{equation*}
\begin{aligned}
\mathbb{E} \left\| \left(\sum_{t=T_0+1}^{T} \boldsymbol{X}^{1}  \boldsymbol{X}^{1 \prime} \right)^{-1} \sum_{t=T_0+1}^{T}\boldsymbol{X}^{1}   \boldsymbol{e}_{t}   \right\|_2^2	& \leq \mathbb{E}\sum_{t=T_0+1}^{T} \left\| \left(\sum_{t=T_0+1}^{T} \boldsymbol{X}^{1}  \boldsymbol{X}^{1 \prime} \right)^{-1} \boldsymbol{X}^{1}  \right\|_2^2\left\|  \boldsymbol{e}_{t}   \right\|_2^2  \\
& =\mathbb{E} \sum_{t=T_0+1}^{T} \left\| \left(T_1 \boldsymbol{X}^{1}  \boldsymbol{X}^{1 \prime} \right)^{-1} \boldsymbol{X}^{1}  \right\|_2^2 \left\|  \boldsymbol{e}_{t}   \right\|_2^2. \\
& =\frac{1}{T_1^{2}} \mathbb{E}\sum_{t=T_0+1}^{T}  \left\| \left( \boldsymbol{X}^{1}  \boldsymbol{X}^{1 \prime} \right)^{-1} \boldsymbol{X}^{1}  \right\|_2^2 \left\|  \boldsymbol{e}_{t}   \right\|_2^2. \\
% & \leq \frac{1}{T_1}   \sum_{t=T_0+1}^{T}\mathbb{E}  \left\| \left( \boldsymbol{X}^{1}  \boldsymbol{X}^{1 \prime} \right)^{-1} \boldsymbol{X}^{1}  \right\|_2 \textcolor{blue}{\sqrt{m} \max_{1\leq i \leq m}{|e_{i,t}|}}.
\end{aligned}
\end{equation*}
Note that by Equation \eqref{eq:trace}, we have
\begin{equation*}
\left\| \left( \boldsymbol{X}^{1}  \boldsymbol{X}^{1 \prime} \right)^{-1} \boldsymbol{X}^{1}  \right\|_2^2 = tr\left\{ \boldsymbol{X}^{1 \prime} \left( \boldsymbol{X}^{1}  \boldsymbol{X}^{1 \prime} \right)^{-1} \left( \boldsymbol{X}^{1}  \boldsymbol{X}^{1 \prime} \right)^{-1} \boldsymbol{X}^{1}  \right\} \leq \frac{d}{m \lambda_1}.
\end{equation*}
Therefore we have
\begin{equation*}
 \mathbb{E} \left\| \left(\sum_{t=T_0+1}^{T} \boldsymbol{X}^{1}  \boldsymbol{X}^{1 \prime} \right)^{-1} \sum_{t=T_0+1}^{T}\boldsymbol{X}^{1}   \boldsymbol{e}_{t}   \right\|_2	 \leq   \frac{d}{m \lambda_1 T_1^2}\mathbb{E}   \sum_{t=T_0+1}^{T}\left\|   \boldsymbol{e}_{t}   \right\|_2^2 =\frac{d}{m \lambda_1 T_1^2}\mathbb{E}   \sum_{t=T_0+1}^{T}\sum_{i=1}^{m} e_{i,t}^2 =\frac{d}{m \lambda_1 T_1} \sum_{i=1}^{m}\mathbb{E}  \frac{1}{T_1} \sum_{t=T_0+1}^{T} e_{i,t}^2 .
\end{equation*}
And by Lemma \ref{lm:sc},
\begin{equation*}
\frac{1}{T_1} \sum_{t=T_0+1}^T e_{i,t}^2 =\frac{1}{T_1} \sum_{t=T_0+1}^T  O_p(T_0^{-1}) = O_p(T_0^{-1}),
\end{equation*}
therefore, with probability $1-o(1)$, there exist a constant $M$ such that $ \frac{1}{T_1} \sum_{t=T_0+1}^{T} e_{i,t}^2  \leq\frac{M }{T_0} $ for any $i \leq m$ and $T_{0}+1 \leq t \leq T$, thus we have 
\begin{eqnarray} \label{eq:term2}
\begin{aligned}
 \mathbb{E} \left\| \left(\sum_{t=T_0+1}^{T} \boldsymbol{X}^{1}  \boldsymbol{X}^{1 \prime} \right)^{-1} \sum_{t=T_0+1}^{T}\boldsymbol{X}^{1}   \boldsymbol{e}_{t}   \right\|_2	 & \leq   \frac{d}{m \lambda_1  T_1} \sum_{i=1}^{m} \frac{M }{T_0} =  \frac{dM}{\lambda_1 T_0  T_1} 
\end{aligned}
\end{eqnarray}
Thus, plug in Equation \eqref{eq:term1} and Equation \eqref{eq:term2}
 into  Equation \eqref{eq:term}, we have
 \begin{equation} \label{eq:CRofTau0}
\begin{aligned}
\mathbb{E} \left[ \left\{\hat{\tau}_1(\boldsymbol{x})- \tau(\boldsymbol{x}) \right\}^2 \mid \boldsymbol{x}  \right] & \leq\|\boldsymbol{x}\|_{2}\left[ \frac{d \left(\sigma^2 + s^2 \right)}{m \lambda_1  T_1}  + \frac{dM}{\lambda_1  T_1 T_0}\right]   =  \frac{d}{\lambda_1}\|\boldsymbol{x}\|_{2} \left[ \frac{\sigma^2 + s^2}{m  T_1}  +  \frac{M}{  T_0 T_1 }\right],
\end{aligned}
 \end{equation}
hold with probability $1-o(1)$.

\textbf{Now, we are able to prove the convergence rate  for H2SXL.}

For the HTE estimator obtained by Algorithm \ref{algo_scx}, we decomposite $\widehat{\tau}(\boldsymbol{x})  - \tau(\boldsymbol{x}) $ into four terms:
\begin{equation*}
\begin{aligned}
 &\widehat{\tau}(\boldsymbol{x})  - \tau(\boldsymbol{x})  = \widehat{e}(\boldsymbol{x}) \widehat{\tau}_{0}(\boldsymbol{x})+\left\{ 1-\widehat{e}(\boldsymbol{x}) \right\}  \widehat{\tau}_{1}(\boldsymbol{x}) - \tau(\boldsymbol{x})\\
& = \widehat{e}(\boldsymbol{x}) \left\{ \widehat{\tau}_{0}(\boldsymbol{x})- \tau(\boldsymbol{x}) \right\} +(1-\widehat{e}(\boldsymbol{x})) \left\{ \widehat{\tau}_{1}(\boldsymbol{x})- \tau(\boldsymbol{x}) \right\} \\
& = \left\{ \widehat{e}(\boldsymbol{x}) - e(\boldsymbol{x}) + e(\boldsymbol{x}) \right\}  \left\{ \widehat{\tau}_{0}(\boldsymbol{x})- \tau(\boldsymbol{x}) \right\} + \left[ 1-e(\boldsymbol{x}) -\left\{ \widehat{e}(\boldsymbol{x}) - e(\boldsymbol{x}) \right\} \right] \left\{ \widehat{\tau}_{1}(\boldsymbol{x})- \tau(\boldsymbol{x}) \right\} \\
& = e(\boldsymbol{x})  \left\{ \widehat{\tau}_{0}(\boldsymbol{x})- \tau(\boldsymbol{x}) \right\}+\left\{ 1-e(\boldsymbol{x})  \right\}   \left\{ \widehat{\tau}_{1}(\boldsymbol{x})- \tau(\boldsymbol{x}) \right\}\\
& +  \left\{ \widehat{e}(\boldsymbol{x}) - e(\boldsymbol{x})\right\} \left\{ \widehat{\tau}_{0}(\boldsymbol{x})- \tau(\boldsymbol{x}) \right\} -  \left\{ \widehat{e}(\boldsymbol{x}) - e(\boldsymbol{x})\right\}  \left\{ \widehat{\tau}_{1}(\boldsymbol{x})- \tau(\boldsymbol{x}) \right\}.
\end{aligned}
\end{equation*}
The assumption $ \widehat{e}(\boldsymbol{x}) - e(\boldsymbol{x}) = o_{\mathbb{P}}(1)$ further implies
\begin{equation*}
\begin{aligned}
  \widehat{\tau}(\boldsymbol{x})  - \tau(\boldsymbol{x})  & = \left\{  o_{\mathbb{P}}(1) + e(\boldsymbol{x}) \right\}  \left\{ \widehat{\tau}_{0}(\boldsymbol{x})- \tau(\boldsymbol{x}) \right\} + \left\{ 1-e(\boldsymbol{x}) + o_{\mathbb{P}}(1) \right\} \left\{ \widehat{\tau}_{1}(\boldsymbol{x})- \tau(\boldsymbol{x}) \right\} \\
\end{aligned}
\end{equation*}
Thus given $\boldsymbol{x} $,
\begin{equation*}
\begin{aligned}
\mathbb{E} \left[ \left\{\hat{\tau}(\boldsymbol{x})- \tau(\boldsymbol{x}) \right\}^2 \mid \boldsymbol{x}  \right] & \leq 2 \mathbb{E} \left( \left\{  o_{\mathbb{P}}(1) + e(\boldsymbol{x}) \right\}^2  \left\{ \widehat{\tau}_{0}(\boldsymbol{x})- \tau(\boldsymbol{x}) \right\}^2 \mid \boldsymbol{x} \right) +  2\mathbb{E} \left( \left\{ 1-e(\boldsymbol{x}) + o_{\mathbb{P}}(1) \right\}^2 \left\{ \widehat{\tau}_{1}(\boldsymbol{x})- \tau(\boldsymbol{x}) \right\}^2 \mid \boldsymbol{x} \right) \\
 & \leq  2\mathbb{E} \left[ \left\{\hat{\tau}_0(\boldsymbol{x})- \tau(\boldsymbol{x}) \right\}^2 \mid \boldsymbol{x}  \right]   +  2\mathbb{E} \left[ \left\{\hat{\tau}_1(\boldsymbol{x})- \tau(\boldsymbol{x}) \right\}^2 \mid \boldsymbol{x}  \right]  .\\
\end{aligned}
\end{equation*}
Recall that by Lemma \ref{lemma:oneside}, we have
\begin{equation*}
\mathbb{E} \left[ \left\{\hat{\tau}_1(\boldsymbol{x})- \tau(\boldsymbol{x}) \right\}^2 \mid \boldsymbol{x}  \right]  =  \mathcal{O}_p\left(  \left( m^{-1} + T_0^{-1}  \right) T_1^{-1}\right).
\end{equation*}
Similarly, we have
\begin{equation*}
\mathbb{E} \left[ \left\{\hat{\tau}_0(\boldsymbol{x})- \tau(\boldsymbol{x}) \right\}^2 \mid \boldsymbol{x}  \right]  =  \mathcal{O}_p\left( \left( n^{-1} + T_0^{-1}  \right) T_1^{-1}\right).
\end{equation*}
Thus
\begin{equation*}
\begin{aligned}
\mathbb{E} \left[ \left\{\hat{\tau}(\boldsymbol{x})- \tau(\boldsymbol{x}) \right\}^2 \mid \boldsymbol{x}  \right] =  \mathcal{O}_p\left( \left( m^{-1} +n^{-1} +   T_0^{-1}  \right) T_1^{-1}   \right).
\end{aligned}
\end{equation*}
Then the proof is completed.

\subsection{Proof for Theorem \ref{thm:H2SDRL}}
In this section, we aim to prove Theorem \ref{thm:H2SDRL}.  Define $\tilde{\tau}(\boldsymbol{X}_i) = \Mean\left( \tilde{\Delta}_{i,T}^{D_{i,T}} \mid \boldsymbol{X}_i \right)$, we have
\begin{equation*}
\phi(\widehat{O}_{i,t})   = (D_{i,T}-\frac{1}{2}){D_{i,T} - e(\boldsymbol{X}_i)\over {e(\boldsymbol{X}_i)\{1-e(\boldsymbol{X}_i)\}}} \left\{\tilde{\Delta}_{i,T}^{D_{i,T}} - \tilde{\tau}(\boldsymbol{X}_i)\right\} +\tilde{\tau}(\boldsymbol{X}_i).
\end{equation*}
Our proof is based on the Proposition 1 in \cite{kennedy2020optimal}  and the proof can be divided into four steps:
\begin{itemize}
\item \textbf{Step 1: }Prove that  the regression estimator $\widehat{\mathbb{E}}_n\{\widehat{\phi}(\widehat{O}_{i,t}) \mid \boldsymbol{X}=\boldsymbol{x}\}$ is stable with respect to distance \begin{equation*} d(\widehat{\phi}, \phi)=\|\widehat{\phi}-\phi\|_{h^2} \equiv \sum_{i,t}\left\{\frac{h_{i,t}\left(\boldsymbol{x}_i\right)^2}{\sum_{i',t'} h_{i',t'}\left(\boldsymbol{x}_{i'} \right)^2}\right\} \int\{\widehat{\phi}(\widehat{O}_{i,t})-\phi(\widehat{O}_{i,t})\}^2 d \mathbb{P}\left(\widehat{O}_{i,t} \mid \boldsymbol{X}_i=\boldsymbol{x}_i\right) \end{equation*}  in the sense of Definition 1 in \cite{kennedy2020optimal}. 
\item \textbf{Step 2: } Calculate the the conditional bias of the estimator $\widehat{\phi}$, i.e., $\widehat{b}(\boldsymbol{x})= \mathbb{E}\left\{\widehat{\phi}(\widehat{O}_{i,t})-\phi(\widehat{O}_{i,t}) \mid  \boldsymbol{X}_i=\boldsymbol{x}\right\} = \frac{1}{2}   \left\{     \frac{1}{1- \widehat{e}(\boldsymbol{x})} -  \frac{1}{ \widehat{e}(\boldsymbol{x})}\right\}   \left\{ \widehat{e}(\boldsymbol{x})-e(\boldsymbol{x}) \right\}\left\{ \tilde{\tau}(\boldsymbol{x})   - \widehat{\tau}(\boldsymbol{x}) \right\}$.
\item \textbf{Step 3: } Prove that  $d(\widehat{\phi}, \phi) \stackrel{p}{\rightarrow} 0$ when  $ \widehat{e}(\boldsymbol{x}) - e(\boldsymbol{x} )   \stackrel{p}{\rightarrow} 0 $ for any $\boldsymbol{x}.$
%\item  \textcolor{blue}{\textbf{Step 4: } Prove that  $\mathbb{E}\{ \phi(\widehat{O}_{i,t}) \mid \boldsymbol{X}_i = \boldsymbol{x}\} = \tau(\boldsymbol{x}) $ .}
\item \textbf{Step 4: } Summarize and prove Theorem \ref{thm:H2SDRL}.
\end{itemize}

\textbf{Step1: Prove that  the regression estimator $\widehat{\mathbb{E}}_n\{\widehat{\phi}(\widehat{O}_{i,t}) \mid \boldsymbol{X}=\boldsymbol{x}\}$ is stable with respect to distance.
\begin{equation*}
d(\widehat{\phi}, \phi)=\|\widehat{\phi}-\phi\|_{h^2} \equiv \sum_{i,t}\left\{\frac{h_{i,t}\left(\boldsymbol{x}_i\right)^2}{\sum_{i',t'} h_{i',t'}\left(\boldsymbol{x}_{i'} \right)^2}\right\} \int\{\widehat{\phi}(\widehat{O}_{i,t})-\phi(\widehat{O}_{i,t})\}^2 d \mathbb{P}\left(\widehat{O}_{i,t} \mid \boldsymbol{X}_i=\boldsymbol{x}_i\right)
\end{equation*} 
in the sense of Definition 1 in \cite{kennedy2020optimal}. } 

By Theorem 1 in \cite{kennedy2020optimal}, it suffices to show that $\operatorname{var}\{\phi(O) \mid \boldsymbol{X}_i = \boldsymbol{x}_i\}$ is bounded.  Note that
\begin{equation*}
\begin{aligned}
\operatorname{var}\{\phi(\widehat{O}_{i,t})  \mid  \boldsymbol{X}_i=\boldsymbol{x}_i \} & = \operatorname{var}\left[(D_{i,T}-\frac{1}{2}) {D_{i,T} - e(\boldsymbol{X}_i)\over {e(\boldsymbol{X}_i)\{1-e(\boldsymbol{X}_i)\}}} \left\{\tilde{\Delta}_{i,T}^{D_{i,T}} - \tilde{\tau}(\boldsymbol{X}_i)\right\} +\tilde{\tau}(\boldsymbol{X}_i)\mid  \boldsymbol{X}_i=\boldsymbol{x}_i \right] \\
& = \operatorname{var}\left[(D_{i,T}-\frac{1}{2}) {D_{i,T} - e(\boldsymbol{X}_i)\over {e(\boldsymbol{X}_i)\{1-e(\boldsymbol{X}_i)\}}} \left\{\tilde{\Delta}_{i,T}^{D_{i,T}} - \tilde{\tau}(\boldsymbol{X}_i)\right\} \mid  \boldsymbol{X}_i=\boldsymbol{x}_i \right] \\
& = \underbrace{\Mean \left ( \operatorname{var}\left[(D_{i,T}-\frac{1}{2}) {D_{i,T} - e(\boldsymbol{X}_i)\over {e(\boldsymbol{X}_i)\{1-e(\boldsymbol{X}_i)\}}} \left\{\tilde{\Delta}_{i,T}^{D_{i,T}} - \tilde{\tau}(\boldsymbol{X}_i)\right\} \mid  \boldsymbol{X}_i=\boldsymbol{x}_i, D_{i,T} \right] \right)} _{\eta_1}\\
& +  \underbrace{\operatorname{var}\left (  \Mean \left[(D_{i,T}-\frac{1}{2}) {D_{i,T} - e(\boldsymbol{X}_i)\over {e(\boldsymbol{X}_i)\{1-e(\boldsymbol{X}_i)\}}} \left\{\tilde{\Delta}_{i,T}^{D_{i,T}} - \tilde{\tau}(\boldsymbol{X}_i)\right\} \mid  \boldsymbol{X}_i=\boldsymbol{x}_i, D_{i,T} \right] \right)} _{\eta_2} \\
\end{aligned}
\end{equation*}
For the second term $\eta_2$, since $\Mean \left( \tilde{\Delta}_{i,T}^{D_{i,T}}\mid  \boldsymbol{X}_i=\boldsymbol{x}_i, D_{i,T}  \right) = \tilde{\tau}(\boldsymbol{X}_i)$ we have
\begin{equation} \label{eq:eta2}
\begin{aligned}
\eta_2 & = \operatorname{var}\left (  \Mean \left[ (D_{i,T}-\frac{1}{2}){D_{i,T} - e(\boldsymbol{X}_i)\over {e(\boldsymbol{X}_i)\{1-e(\boldsymbol{X}_i)\}}} \left\{\tilde{\Delta}_{i,T}^{D_{i,T}} - \tilde{\tau}(\boldsymbol{X}_i)\right\} \mid  \boldsymbol{X}_i=\boldsymbol{x}_i, D_{i,T} \right] \right) \\
& = \operatorname{var}\left ( (D_{i,T}-\frac{1}{2}) {D_{i,T} - e(\boldsymbol{X}_i)\over {e(\boldsymbol{X}_i)\{1-e(\boldsymbol{X}_i)\}}} \left[  \Mean \left\{ \tilde{\Delta}_{i,T}^{D_{i,T}} \right\}- \tilde{\tau}(\boldsymbol{X}_i)\right] \mid  \boldsymbol{X}_i=\boldsymbol{x}_i, D_{i,T}  \right)\\
& = \operatorname{var}\left ((D_{i,T}-\frac{1}{2})  {D_{i,T} - e(\boldsymbol{x}_i)\over {e(\boldsymbol{x}_i)\{1-e(\boldsymbol{x}_i)\}}} \left[ \tilde{\tau}(\boldsymbol{X}_i)- \tilde{\tau}(\boldsymbol{X}_i)\right] \mid   D_{i,T}  \right)  = 0.
\end{aligned}
\end{equation}

For the first term $\eta_1$, denote 
\begin{equation*}
	U = \max \left\{  \left| 1 \over {e(\boldsymbol{X}_i) } \right| , \left| { 1 \over {1-e(\boldsymbol{X}_i)} }\right| \right\},
\end{equation*}
then we have $U < \frac{1}{c} < \infty$, 
since  $e(\boldsymbol{X}_i) >c $ and $1-e(\boldsymbol{X}_i) <1- c $ by assumption \ref{ass:Positivity}. Therefore,
\begin{equation} \label{eq:eta1}
\begin{aligned}
 \eta_1 & \leq  \Mean \left ( \operatorname{var}\left[ 	\frac{U}{2}  \left\{\tilde{\Delta}_{i,T}^{D_{i,T}} - \tilde{\tau}(\boldsymbol{X}_i)\right\} \mid  \boldsymbol{X}_i=\boldsymbol{x}_i, D_{i,T} \right] \right) = 	\frac{U^2}{4}\Mean \left [ \operatorname{var} \left\{\tilde{\Delta}_{i,T}^{D_{i,T}}  \right\} \mid  \boldsymbol{X}_i=\boldsymbol{x}_i, D_{i,T} \right] \\
 & < \frac{1}{4c^2} \Mean \left [ \operatorname{var} \left\{\tilde{\Delta}_{i,T}^{D_{i,T}}  \right\} \mid  \boldsymbol{X}_i=\boldsymbol{x}_i, D_{i,T} \right].
\end{aligned}
 \end{equation}
By combining the Equation \eqref{eq:eta1} and Equation \eqref{eq:eta2}, $\operatorname{var}\{\phi(\widehat{O}_{i,t})  \mid  \boldsymbol{X}_i=\boldsymbol{x}_i \} $ is bounded by
\begin{equation*}
\operatorname{var}\{\phi(\widehat{O}_{i,t})  \mid  \boldsymbol{X}_i=\boldsymbol{x}_i \}   =  \eta_1 +  \eta_2 =  \eta_1 < \frac{1}{4c^2}\Mean \left [ \operatorname{var} \left\{\tilde{\Delta}_{i,T}^{D_{i,T}}  \right\} \mid  \boldsymbol{X}_i=\boldsymbol{x}_i, D_{i,T} \right].
\end{equation*}
Thus, to prove $\operatorname{var}\{\phi(\widehat{O}_{i,t})  \mid  \boldsymbol{X}_i=\boldsymbol{x}_i \}$ is bounded, it suffices to prove that $\Mean \left [ \operatorname{var} \left\{\tilde{\Delta}_{i,T}^{D_{i,T}}  \right\} \mid  \boldsymbol{X}_i=\boldsymbol{x}_i, D_{i,T} \right]$ is bounded. We next focus on proving that $\operatorname{var} \left\{\tilde{\Delta}_{i,T}^{1}\mid  \boldsymbol{X}_i=\boldsymbol{x}_i, D_{i,T}=0   \right\} $ is bounded, and $\operatorname{var} \left\{\tilde{\Delta}_{i,T}^{0}\mid  \boldsymbol{X}_i=\boldsymbol{x}_i, D_{i,T}=1   \right\} $  can be proved similarly.

By Equation \eqref{eq:DeltaDecom}, we have $\tilde{\Delta}_{i,t}^{1} =  \tau(\boldsymbol{x}_i) + \overline{\epsilon}_{i,t} +\varepsilon_{i,t}^{0} +e_{i,t}$. Applying Cauchy-Schwarz inequality, 
\begin{equation*}
\begin{aligned}
\operatorname{var} \left\{\tilde{\Delta}_{i,T}^{1}\mid  \boldsymbol{X}_i=\boldsymbol{x}_i, D_{i,T}=0   \right\}  & = \operatorname{var} \left\{\tau(\boldsymbol{x}_i) + \overline{\epsilon}_{i,t} +\varepsilon_{i,t}^{0} +e_{i,t} \mid  \boldsymbol{X}_i=\boldsymbol{x}_i, D_{i,T}=0   \right\}\\
&  \leq 3   \left[ \operatorname{var} \left\{   \overline{\epsilon}_{i,t}  \right\} +\operatorname{var} \left\{   \varepsilon_{i,t}^{0}  \right\} +\operatorname{var} \left\{  e_{i,t}  \right\} \right]  = 3   \left[ \sigma^2 + s^2 + \operatorname{var} \left\{  e_{i,t}  \right\} \right] \\
\end{aligned}
\end{equation*}
Recall that $e_{i,t} = \Mean\{Y_{i,t}(0) \} - \widehat{Y}_{i,t}(0) = \mathbf{Y}_t^{con \prime} \left(\widehat{\mathbf{w}}_{i} -\mathbf{w}_{i} \right)$, which follows 
\begin{equation} \label{eq:var} \operatorname{var} \left\{  e_{i,t}  \right\}  = \mathbf{Y}_t^{con \prime}  \operatorname{var} \left\{  \widehat{\mathbf{w}}_{i} -\mathbf{w}_{i}\right\} \mathbf{Y}_t^{con }= \mathbf{Y}_t^{con \prime}  \operatorname{var} \left\{  \widehat{\mathbf{w}}_{i} \right\} \mathbf{Y}_t^{con }. 
\end{equation}
Since $\|\widehat{\mathbf{w}}_{i}  \leq K\|$, we have $\operatorname{var} \left\{  \widehat{\mathbf{w}}_{i} \right\}$ is bounded, which follows immediately that $\operatorname{var} \left\{\tilde{\Delta}_{i,T}^{0}\mid  \boldsymbol{X}_i=\boldsymbol{x}_i, D_{i,T}=0   \right\} $ is bounded. Therefore the proof of Step 1 is completed.

\textbf{Step 2: Calculate the the conditional bias of the estimator $\widehat{\phi}$, i.e., $\widehat{b}(\boldsymbol{x})= \mathbb{E}\left\{\widehat{\phi}(\widehat{O}_{i,t})-\phi(\widehat{O}_{i,t}) \mid  \boldsymbol{X}_i=\boldsymbol{x}\right\}$.}

We firstly consider  $\mathbb{E}\left\{ \phi (\widehat{O}_{i,t}) \mid  \boldsymbol{X}_i=\boldsymbol{x}\right\}$.
\begin{equation} \label{eq:Eofphi}
\begin{aligned}
\mathbb{E}\left\{ \phi(\widehat{O}_{i,t})\mid  \boldsymbol{X}_i=\boldsymbol{x}\right\}    & =  \mathbb{E}\left[ {D_{i,T} - e(\boldsymbol{x})\over {e(\boldsymbol{x})\{1-e(\boldsymbol{x})\}}} \left\{\tilde{\Delta}_{i,T}^{D_{i,T}} - \tilde{\tau}(\boldsymbol{x})\right\} +\tilde{\tau}(\boldsymbol{x})\mid  \boldsymbol{X}_i=\boldsymbol{x} \right] \\
& =  \mathbb{E} \left( \mathbb{E}\left[ {D_{i,T} - e(\boldsymbol{X}_i)\over {e(\boldsymbol{X}_i)\{1-e(\boldsymbol{X}_i)\}}} \left\{\tilde{\Delta}_{i,T}^{D_{i,T}} - \tilde{\tau}(\boldsymbol{X}_i)\right\} +\tilde{\tau}(\boldsymbol{X}_i) \mid D_{i,T} \right] \mid  \boldsymbol{X}_i=\boldsymbol{x} \right) \\
& =  \mathbb{E} \left( {D_{i,T} - e(\boldsymbol{X}_i)\over {e(\boldsymbol{X}_i)\{1-e(\boldsymbol{X}_i)\}}} \left[ \mathbb{E} \left\{\tilde{\Delta}_{i,T}^{D_{i,T}} \mid D_{i,T}, \boldsymbol{X}_i=\boldsymbol{x}   \right\}  - \tilde{\tau}(\boldsymbol{X}_i) \right]+\tilde{\tau}(\boldsymbol{X}_i)  \mid  \boldsymbol{X}_i=\boldsymbol{x} \right) \\
& =  \mathbb{E} \left( {D_{i,T} - e(\boldsymbol{X}_i)\over {e(\boldsymbol{X}_i)\{1-e(\boldsymbol{X}_i)\}}} \left\{ \tilde{\tau}(\boldsymbol{X}_i)  - \tilde{\tau}(\boldsymbol{X}_i) \right\}+\tilde{\tau}(\boldsymbol{X}_i)  \mid  \boldsymbol{X}_i=\boldsymbol{x} \right) \\
& = \tilde{\tau}(\boldsymbol{x}) .
\end{aligned}
\end{equation}
Thus
\begin{equation*}
\begin{aligned}
\widehat{b}(\boldsymbol{x}) & = \mathbb{E}\left\{\widehat{\phi}(\widehat{O}_{i,t})-\phi(\widehat{O}_{i,t}) \mid  \boldsymbol{X}_i=\boldsymbol{x}\right\}= \mathbb{E}\left\{\widehat{\phi}(\widehat{O}_{i,t}) \mid  \boldsymbol{X}_i=\boldsymbol{x}\right\}
 -\mathbb{E}\left\{\phi(\widehat{O}_{i,t}) \mid  \boldsymbol{X}_i=\boldsymbol{x}_i\right\} \\
 & = \mathbb{E}\left\{\widehat{\phi}(\widehat{O}_{i,t}) \mid  \boldsymbol{X}_i=\boldsymbol{x}\right\}-\tilde{\tau}(\boldsymbol{x})  = \mathbb{E}\left\{\widehat{\phi}(\widehat{O}_{i,t})-\tilde{\tau}(\boldsymbol{X}_i) 
 \mid  \boldsymbol{X}_i=\boldsymbol{x}\right\} \\
 &  =  \mathbb{E} \left[ \mathbb{E}\left\{ \widehat{\phi}(\widehat{O}_{i,t})-\tilde{\tau}(\boldsymbol{X}_i) \mid D_{i,T} \right\} \mid  \boldsymbol{X}_i=\boldsymbol{x} \right].
 \end{aligned}
\end{equation*}
Then we calculate $ \mathbb{E}\left\{ \widehat{\phi}(\widehat{O}_{i,t})-\tilde{\tau}(\boldsymbol{X}_i) \mid D_{i,T} , \boldsymbol{X}_i=\boldsymbol{x}  \right\} 
$.
\begin{equation*}
\begin{aligned}
&  \mathbb{E}\left\{ \widehat{\phi}(\widehat{O}_{i,t})-\tilde{\tau}(\boldsymbol{X}_i) \mid D_{i,T} , \boldsymbol{X}_i=\boldsymbol{x}  \right\}  \\
& =  \mathbb{E}\left[ (D_{i,T}-\frac{1}{2}) {D_{i,T} - \widehat{e}(\boldsymbol{X}_i)\over {\widehat{e}(\boldsymbol{X}_i)\{1-\widehat{e}(\boldsymbol{X}_i)\}}} \left\{\tilde{\Delta}_{i,T}^{D_{i,T}} - \widehat{\tau}(\boldsymbol{X}_i)\right\} + \widehat{\tau}(\boldsymbol{X}_i) -\tilde{\tau}(\boldsymbol{X}_i)\mid D_{i,T},\boldsymbol{X}_i=\boldsymbol{x} \right] \\
& = (D_{i,T}-\frac{1}{2}) {D_{i,T} - \widehat{e}(\boldsymbol{x})\over {\widehat{e}(\boldsymbol{x})\{1-\widehat{e}(\boldsymbol{x})\}}} \left[ \mathbb{E} \left\{\tilde{\Delta}_{i,T}^{D_{i,T}} \mid D_{i,T}, \boldsymbol{X}_i=\boldsymbol{x}   \right\}  - \widehat{\tau}(\boldsymbol{x}) \right]+\widehat{\tau}(\boldsymbol{x}) -\tilde{\tau}(\boldsymbol{x})  \\
& = (D_{i,T}-\frac{1}{2}) {D_{i,T} - \widehat{e}(\boldsymbol{x})\over {\widehat{e}(\boldsymbol{x})\{1-\widehat{e}(\boldsymbol{x})\}}} \left\{ \tilde{\tau}(\boldsymbol{x})   - \widehat{\tau}(\boldsymbol{x}) \right\}+\widehat{\tau}(\boldsymbol{x}) -\tilde{\tau}(\boldsymbol{x})\\
& = \left\{(D_{i,T}-\frac{1}{2}) {D_{i,T} - \widehat{e}(\boldsymbol{x})\over {\widehat{e}(\boldsymbol{x})\{1-\widehat{e}(\boldsymbol{x})\}}}   -1 \right\} \left\{ \tilde{\tau}(\boldsymbol{x})   - \widehat{\tau}(\boldsymbol{x}) \right\} . 
\end{aligned}
\end{equation*}
Therefore, 
\begin{equation*}
\begin{aligned}
\widehat{b}(\boldsymbol{x}) &  =  \mathbb{E} \left[ \mathbb{E}\left\{ \widehat{\phi}(\widehat{O}_{i,t})-\tilde{\tau}(\boldsymbol{X}_i) \mid D_{i,T} \right\} \mid  \boldsymbol{X}_i=\boldsymbol{x} \right]\\
& =  \mathbb{E}_{D_{i,T}} \left[  \left\{(D_{i,T}-\frac{1}{2}) {D_{i,T} - \widehat{e}(\boldsymbol{x})\over {\widehat{e}(\boldsymbol{x})\{1-\widehat{e}(\boldsymbol{x})\}}}   -1 \right\} \left\{ \tilde{\tau}(\boldsymbol{x})   - \widehat{\tau}(\boldsymbol{x}) \right\}  \right] \\
& =  \left[\mathbb{E}_{D_{i,T}}   \left\{(D_{i,T}-\frac{1}{2}) {D_{i,T} - \widehat{e}(\boldsymbol{x})\over {\widehat{e}(\boldsymbol{x})\{1-\widehat{e}(\boldsymbol{x})\}}} \right\}   -1 \right] \left\{ \tilde{\tau}(\boldsymbol{x})   - \widehat{\tau}(\boldsymbol{x}) \right\}  \ \\
& =  \left[\mathbb{E}_{D_{i,T}}   \left\{   \frac{1}{2} \frac{D_{i,T}}{ \widehat{e}(\boldsymbol{x})}+  \frac{1}{2} \frac{1-D_{i,T}}{1- \widehat{e}(\boldsymbol{x})}\right\} -1  \right] \left\{ \tilde{\tau}(\boldsymbol{x})   - \widehat{\tau}(\boldsymbol{x}) \right\}  \\
\end{aligned}
\end{equation*}
Note that $\Mean\left(D_{i,T}  \mid  \boldsymbol{X}_i=\boldsymbol{x}\right) = e(\boldsymbol{x})$, thus we have
\begin{equation*}
\begin{aligned}
\widehat{b}(\boldsymbol{x})  & =  \left[\frac{1}{2}   \left\{    \frac{e(\boldsymbol{x})}{ \widehat{e}(\boldsymbol{x})}+   \frac{1-e(\boldsymbol{x})}{1- \widehat{e}(\boldsymbol{x})}\right\} -1  \right] \left\{ \tilde{\tau}(\boldsymbol{x})   - \widehat{\tau}(\boldsymbol{x}) \right\}  \ \\
& =  \frac{1}{2}   \left\{     \frac{1}{1- \widehat{e}(\boldsymbol{x})} -  \frac{1}{ \widehat{e}(\boldsymbol{x})}\right\}   \left\{ \widehat{e}(\boldsymbol{x})-e(\boldsymbol{x}) \right\}\left\{ \tilde{\tau}(\boldsymbol{x})   - \widehat{\tau}(\boldsymbol{x}) \right\}.  \ \\ 
\end{aligned}
\end{equation*}

\textbf{Step 3: Prove that  $d(\widehat{\phi}, \phi) \stackrel{p}{\rightarrow} 0$ when  $ \widehat{e}(\boldsymbol{x}) - e(\boldsymbol{x} )   \stackrel{p}{\rightarrow} 0 $ for any $\boldsymbol{x}$.}

By definition ,
\begin{equation} \label{eq:d}
\begin{aligned}
d(\widehat{\phi}, \phi)& =  \sum_{i,t}\left\{\frac{h_{i,t}\left(\boldsymbol{x}_i\right)^2}{\sum_{i',t'} h_{i',t'}\left(\boldsymbol{x}_{i'} \right)^2}\right\} \int\{\widehat{\phi}(\widehat{O}_{i,t})-\phi(\widehat{O}_{i,t})\}^2 d \mathbb{P}\left(\widehat{O}_{i,t} \mid \boldsymbol{X}_i=\boldsymbol{x}_i\right)  \\
&=  \sum_{i,t}\left\{\frac{h_{i,t}\left(\boldsymbol{x}_i\right)^2}{\sum_{i',t'} h_{i',t'}\left(\boldsymbol{x}_{i'} \right)^2}\right\} \Mean \left[ \left\{\widehat{\phi}(\widehat{O}_{i,t})-\phi(\widehat{O}_{i,t})\right\}^2 \mid \boldsymbol{X}_i=\boldsymbol{x}_i\right].  \\
\end{aligned}
\end{equation}
We firstly calculate  $\widehat{\phi}(\widehat{O}_{i,t})-\phi(\widehat{O}_{i,t})  $.
% Thus it suffices to show that $\widehat{\phi}(\widehat{O}_{i,t})-\phi(\widehat{O}_{i,t})  \stackrel{p}{\rightarrow} 0 $ by Continuous mapping theorem.
\begin{equation*}
\begin{aligned}
 \widehat{\phi}(\widehat{O}_{i,t})-\phi(\widehat{O}_{i,t})  
  & =   \widehat{\phi}(\widehat{O}_{i,t})  -  (D_{i,T}-\frac{1}{2}) {D_{i,T} - \widehat{e}(\boldsymbol{X}_i)\over {\widehat{e}(\boldsymbol{X}_i)\{1-\widehat{e}(\boldsymbol{X}_i)\}}} \left\{\tilde{\Delta}_{i,T}^{D_{i,T}} - \tilde{\tau}(\boldsymbol{X}_i)\right\}\\
  & + (D_{i,T}-\frac{1}{2}) {D_{i,T} - \widehat{e}(\boldsymbol{X}_i)\over {\widehat{e}(\boldsymbol{X}_i)\{1-\widehat{e}(\boldsymbol{X}_i)\}}} \left\{\tilde{\Delta}_{i,T}^{D_{i,T}} - \tilde{\tau}(\boldsymbol{X}_i)\right\} - \phi(\widehat{O}_{i,t})  \\
   & =(D_{i,T}-\frac{1}{2}) {D_{i,T} - \widehat{e}(\boldsymbol{X}_i)\over {\widehat{e}(\boldsymbol{X}_i)\{1-\widehat{e}(\boldsymbol{X}_i)\}}} \left\{\tilde{\tau}(\boldsymbol{X}_i) - \widehat{\tau}(\boldsymbol{X}_i)\right\}  +\widehat{\tau}(\boldsymbol{X}_i) \\
  & + (D_{i,T}-\frac{1}{2}) \left[ {D_{i,T} - \widehat{e}(\boldsymbol{X}_i)\over {\widehat{e}(\boldsymbol{X}_i)\{1-\widehat{e}(\boldsymbol{X}_i)\}}} - {D_{i,T} - e(\boldsymbol{X}_i)\over {e(\boldsymbol{X}_i)\{1-e(\boldsymbol{X}_i)\}}}\right]\left\{\tilde{\Delta}_{i,T}^{D_{i,T}} - \tilde{\tau}(\boldsymbol{X}_i)\right\} -\tilde{\tau}(\boldsymbol{X}_i)\\
     & =   \left[ (D_{i,T}-\frac{1}{2}) {D_{i,T} - \widehat{e}(\boldsymbol{X}_i)\over {\widehat{e}(\boldsymbol{X}_i)\{1-\widehat{e}(\boldsymbol{X}_i)\}}}  -1\right] \left\{\tilde{\tau}(\boldsymbol{X}_i) - \widehat{\tau}(\boldsymbol{X}_i)\right\}  \\
  & + (D_{i,T}-\frac{1}{2}) \left[ {D_{i,T} - \widehat{e}(\boldsymbol{X}_i)\over {\widehat{e}(\boldsymbol{X}_i)\{1-\widehat{e}(\boldsymbol{X}_i)\}}} - {D_{i,T} - e(\boldsymbol{X}_i)\over {e(\boldsymbol{X}_i)\{1-e(\boldsymbol{X}_i)\}}}\right]\left\{\tilde{\Delta}_{i,T}^{D_{i,T}} - \tilde{\tau}(\boldsymbol{X}_i)\right\} \\
\end{aligned}
\end{equation*}
Note that
\begin{equation*}
(D_{i,T}-\frac{1}{2}) {D_{i,T} - \widehat{e}(\boldsymbol{X}_i)\over {\widehat{e}(\boldsymbol{X}_i)\{1-\widehat{e}(\boldsymbol{X}_i)\}}}  -1 = \frac{D_{i,T} }{2\widehat{e}(\boldsymbol{X}_i)}+\frac{1-D_{i,T} }{2 \{ 1-\widehat{e}(\boldsymbol{X}_i)\}}-1 = \frac{1}{2}	 \left\{  \frac{1}{1-\widehat{e}(\boldsymbol{X}_i)} - \frac{1}{\widehat{e}(\boldsymbol{X}_i)}\right\} \left\{\widehat{e}(\boldsymbol{X}_i)-D_{i,T} \right\},
\end{equation*}
and
\begin{equation*}
{D_{i,T} - \widehat{e}(\boldsymbol{X}_i)\over {\widehat{e}(\boldsymbol{X}_i)\{1-\widehat{e}(\boldsymbol{X}_i)\}}} - {D_{i,T} - e(\boldsymbol{X}_i)\over {e(\boldsymbol{X}_i)\{1-e(\boldsymbol{X}_i)\}}}  = \left\{ \widehat{e}(\boldsymbol{X}_i) - e(\boldsymbol{X}_i )\right\}\frac{D_{i,T}\left\{ \widehat{e}(\boldsymbol{X}_i) + e(\boldsymbol{X}_i ) - 1\right\}-\widehat{e}(\boldsymbol{X}_i) e(\boldsymbol{X}_i )}{\widehat{e}(\boldsymbol{X}_i)\{1-\widehat{e}(\boldsymbol{X}_i)\} (\boldsymbol{X}_i)\{1-e(\boldsymbol{X}_i)\}}.
\end{equation*}
Therefore we have
\begin{equation*}
\begin{aligned}
\widehat{\phi}(\widehat{O}_{i,t})-\phi(\widehat{O}_{i,t})   
& =  \underbrace{ \frac{1}{2}	 \left\{  \frac{1}{1-\widehat{e}(\boldsymbol{X}_i)} - \frac{1}{\widehat{e}(\boldsymbol{X}_i)}\right\} }_{g_1(\boldsymbol{X}_i)}\left\{\widehat{e}(\boldsymbol{X}_i)-D_{i,T} \right\}\left\{\tilde{\tau}(\boldsymbol{X}_i) - \widehat{\tau}(\boldsymbol{X}_i)\right\}  \\
  & +  \underbrace{(D_{i,T}-\frac{1}{2})\frac{D_{i,T}\left\{ \widehat{e}(\boldsymbol{X}_i) + e(\boldsymbol{X}_i ) - 1\right\}-\widehat{e}(\boldsymbol{X}_i) e(\boldsymbol{X}_i )}{\widehat{e}(\boldsymbol{X}_i)\{1-\widehat{e}(\boldsymbol{X}_i)\} (\boldsymbol{X}_i)\{1-e(\boldsymbol{X}_i)\}} }_{g_2(D_{i,T},\boldsymbol{X}_i)}
  \left\{ \widehat{e}(\boldsymbol{X}_i) - e(\boldsymbol{X}_i )\right\}\left\{\tilde{\Delta}_{i,T}^{D_{i,T}} - \tilde{\tau}(\boldsymbol{X}_i)\right\} \\
  & = g_1(\boldsymbol{X}_i)\left\{\widehat{e}(\boldsymbol{X}_i)-D_{i,T} \right\}\left\{\tilde{\tau}(\boldsymbol{X}_i) - \widehat{\tau}(\boldsymbol{X}_i)\right\} +g_2(D_{i,T},\boldsymbol{X}_i)
  \left\{ \widehat{e}(\boldsymbol{X}_i) - e(\boldsymbol{X}_i )\right\}\left\{\tilde{\Delta}_{i,T}^{D_{i,T}} - \tilde{\tau}(\boldsymbol{X}_i)\right\}.
\end{aligned}
\end{equation*}
Then we consider $ \Mean \left[ \left\{\widehat{\phi}(\widehat{O}_{i,t})-\phi(\widehat{O}_{i,t})\right\}^2 \mid \boldsymbol{X}_i=\boldsymbol{x}_i\right]$.
\begin{equation*}
\begin{aligned}
&  \Mean \left[ \left\{\widehat{\phi}(\widehat{O}_{i,t})-\phi(\widehat{O}_{i,t})\right\}^2 \mid \boldsymbol{X}_i=\boldsymbol{x}_i\right]\\
 & =  \Mean  \left( \left[ g_1(\boldsymbol{X}_i)\left\{\widehat{e}(\boldsymbol{X}_i)-D_{i,T} \right\}\left\{\tilde{\tau}(\boldsymbol{X}_i) - \widehat{\tau}(\boldsymbol{X}_i)\right\} +g_2(D_{i,T},\boldsymbol{X}_i)
  \left\{ \widehat{e}(\boldsymbol{X}_i) - e(\boldsymbol{X}_i )\right\}\left\{\tilde{\Delta}_{i,T}^{D_{i,T}} - \tilde{\tau}(\boldsymbol{X}_i)\right\} \right]^2\mid \boldsymbol{X}_i=\boldsymbol{x}_i \right) \\
  &  =  \underbrace{\Mean \left[ g_1^2(\boldsymbol{X}_i)\left\{\widehat{e}(\boldsymbol{X}_i)-D_{i,T} \right\}^2\left\{\tilde{\tau}(\boldsymbol{X}_i) - \widehat{\tau}(\boldsymbol{X}_i)\right\}^2 \mid \boldsymbol{X}_i=\boldsymbol{x}_i\right]}_{\psi_1}\\
   &  +\underbrace{ \Mean \left[g_2^2(D_{i,T},\boldsymbol{X}_i)
  \left\{ \widehat{e}(\boldsymbol{X}_i) - e(\boldsymbol{X}_i )\right\}^2\left\{\tilde{\Delta}_{i,T}^{D_{i,T}} - \tilde{\tau}(\boldsymbol{X}_i)\right\}^2\mid \boldsymbol{X}_i=\boldsymbol{x}_i\right]}_{\psi_2}\\
    &  + \underbrace{\Mean \left[ 2 g_1(\boldsymbol{X}_i)\left\{\widehat{e}(\boldsymbol{X}_i)-D_{i,T} \right\}\left\{\tilde{\tau}(\boldsymbol{X}_i) - \widehat{\tau}(\boldsymbol{X}_i)\right\}g_2(D_{i,T},\boldsymbol{X}_i)
  \left\{ \widehat{e}(\boldsymbol{X}_i) - e(\boldsymbol{X}_i )\right\}\left\{\tilde{\Delta}_{i,T}^{D_{i,T}} - \tilde{\tau}(\boldsymbol{X}_i)\right\}  \mid \boldsymbol{X}_i=\boldsymbol{x}_i\right]}_{\psi_3}\\
\end{aligned}
\end{equation*}
We firstly consider $\psi_1 $. 
\begin{equation*}
\begin{aligned}
\psi_1 & = \Mean \left[ g_1^2(\boldsymbol{X}_i)\left\{\widehat{e}(\boldsymbol{X}_i)-D_{i,T} \right\}^2\left\{\tilde{\tau}(\boldsymbol{X}_i) - \widehat{\tau}(\boldsymbol{X}_i)\right\}^2 \mid \boldsymbol{X}_i=\boldsymbol{x}_i\right] \\
 &= g_1^2(\boldsymbol{x}_i) \left\{\tilde{\tau}(\boldsymbol{x}_i) - \widehat{\tau}(\boldsymbol{x}_i)\right\}^2 \Mean \left[ \left\{\widehat{e}(\boldsymbol{X}_i)-D_{i,T} \right\}^2 \mid \boldsymbol{X}_i=\boldsymbol{x}_i\right] \\
  &= g_1^2(\boldsymbol{x}_i) \left\{\tilde{\tau}(\boldsymbol{x}_i) - \widehat{\tau}(\boldsymbol{x}_i)\right\}^2 \left[  e(\boldsymbol{x}_i) \left\{\widehat{e}(\boldsymbol{x}_i)-1 \right\}^2   + \left\{1-e(\boldsymbol{x}_i)  \right\}\widehat{e}^2(\boldsymbol{x}_i)
  \right] \\
    &= g_1^2(\boldsymbol{x}_i) \left\{\tilde{\tau}(\boldsymbol{x}_i) - \widehat{\tau}(\boldsymbol{x}_i)\right\}^2 \left\{  e(\boldsymbol{x}_i) -\widehat{e}(\boldsymbol{x}_i) \right\}^2  .
\end{aligned}
\end{equation*}
Then consider $\psi_2 $. 
\begin{equation*}
\begin{aligned}
\psi_2 & = \Mean \left[g_2^2(D_{i,T},\boldsymbol{X}_i)
  \left\{ \widehat{e}(\boldsymbol{X}_i) - e(\boldsymbol{X}_i )\right\}^2\left\{\tilde{\Delta}_{i,T}^{D_{i,T}} - \tilde{\tau}(\boldsymbol{X}_i)\right\}^2\mid \boldsymbol{X}_i=\boldsymbol{x}_i\right] \\
  & = \Mean \left(\Mean \left[g_2^2(D_{i,T},\boldsymbol{X}_i)
  \left\{ \widehat{e}(\boldsymbol{X}_i) - e(\boldsymbol{X}_i )\right\}^2\left\{\tilde{\Delta}_{i,T}^{D_{i,T}} - \tilde{\tau}(\boldsymbol{X}_i)\right\}^2\mid D_{i,T} \right]\mid \boldsymbol{X}_i=\boldsymbol{x}_i\right) \\
  & =\left\{ \widehat{e}(\boldsymbol{x}_i) - e(\boldsymbol{x}_i )\right\}^2 \Mean\left(  g_2^2(D_{i,T},\boldsymbol{x}_i) \Mean \left[
  \left\{\tilde{\Delta}_{i,T}^{D_{i,T}} - \tilde{\tau}(\boldsymbol{x}_i)\right\}^2 \mid D_{i,T} \right]\mid \boldsymbol{X}_i=\boldsymbol{x}_i\right) \\
      & \triangleq \left\{ \widehat{e}(\boldsymbol{x}_i) - e(\boldsymbol{x}_i )\right\}^2 f_1(D_{i,T},\boldsymbol{X}_i).
\end{aligned}
\end{equation*}
Lastly, we focus on  $\psi_3 $. 
\begin{equation*}
\begin{aligned}
\psi_3 & = \Mean \left[ 2 g_1(\boldsymbol{X}_i)\left\{\widehat{e}(\boldsymbol{X}_i)-D_{i,T} \right\}\left\{\tilde{\tau}(\boldsymbol{X}_i) - \widehat{\tau}(\boldsymbol{X}_i)\right\}g_2(D_{i,T},\boldsymbol{X}_i)
  \left\{ \widehat{e}(\boldsymbol{X}_i) - e(\boldsymbol{X}_i )\right\}\left\{\tilde{\Delta}_{i,T}^{D_{i,T}} - \tilde{\tau}(\boldsymbol{X}_i)\right\}  \mid \boldsymbol{X}_i=\boldsymbol{x}_i\right] \\
 & = \left\{ \widehat{e}(\boldsymbol{x}_i) - e(\boldsymbol{x}_i )\right\} \left\{\tilde{\tau}(\boldsymbol{x}_i) - \widehat{\tau}(\boldsymbol{x}_i)\right\} 2 g_1(\boldsymbol{x}_i)  \Mean \left[ \left\{\widehat{e}(\boldsymbol{X}_i)-D_{i,T} \right\}g_2(D_{i,T},\boldsymbol{X}_i)
  \left\{\tilde{\Delta}_{i,T}^{D_{i,T}} - \tilde{\tau}(\boldsymbol{X}_i)\right\}  \mid \boldsymbol{X}_i=\boldsymbol{x}_i\right] \\
& \triangleq \left\{ \widehat{e}(\boldsymbol{x}_i) - e(\boldsymbol{x}_i )\right\} \left\{\tilde{\tau}(\boldsymbol{x}_i) - \widehat{\tau}(\boldsymbol{x}_i)\right\}  f_2(D_{i,T},\boldsymbol{X}_i).
\end{aligned}
\end{equation*}
Hence,
\begin{equation*}
\begin{aligned}
& \Mean \left[ \left\{\widehat{\phi}(\widehat{O}_{i,t})-\phi(\widehat{O}_{i,t})\right\}^2 \mid \boldsymbol{X}_i=\boldsymbol{x}_i\right]   = \psi_1 + \psi_2 +\psi_3 \\
&  =  g_1^2(\boldsymbol{x}_i) \left\{\tilde{\tau}(\boldsymbol{x}_i) - \widehat{\tau}(\boldsymbol{x}_i)\right\}^2 \left\{  e(\boldsymbol{x}_i) -\widehat{e}(\boldsymbol{x}_i) \right\}^2 + \left\{ \widehat{e}(\boldsymbol{x}_i) - e(\boldsymbol{x}_i )\right\}^2 f_1(D_{i,T},\boldsymbol{X}_i)\left\{ \widehat{e}(\boldsymbol{x}_i) - e(\boldsymbol{x}_i )\right\} \left\{\tilde{\tau}(\boldsymbol{x}_i) - \widehat{\tau}(\boldsymbol{x}_i)\right\}  f_2(D_{i,T},\boldsymbol{X}_i) \\
& =\left\{  e(\boldsymbol{x}_i) -\widehat{e}(\boldsymbol{x}_i) \right\} \left[ g_1^2(\boldsymbol{x}_i) \left\{\tilde{\tau}(\boldsymbol{x}_i) - \widehat{\tau}(\boldsymbol{x}_i)\right\}^2 \left\{  e(\boldsymbol{x}_i) -\widehat{e}(\boldsymbol{x}_i) \right\} + \left\{ \widehat{e}(\boldsymbol{x}_i) - e(\boldsymbol{x}_i )\right\} f_1(D_{i,T},\boldsymbol{X}_i) \left\{\tilde{\tau}(\boldsymbol{x}_i) - \widehat{\tau}(\boldsymbol{x}_i)\right\}  f_2(D_{i,T},\boldsymbol{X}_i)  \right]
\end{aligned}
\end{equation*}
Therefore, $\Mean \left[ \left\{\widehat{\phi}(\widehat{O}_{i,t})-\phi(\widehat{O}_{i,t})\right\}^2 \mid \boldsymbol{X}_i=\boldsymbol{x}_i\right]   \stackrel{p}{\rightarrow} 0 $ when  $ \widehat{e}(\boldsymbol{x}_i) - e(\boldsymbol{x}_i )   \stackrel{p}{\rightarrow} 0 $ for any $\boldsymbol{x}_i$.  Then by Equation \eqref{eq:d},
\begin{equation*}
d(\widehat{\phi}, \phi) =  \sum_{i,t}\left\{\frac{h_{i,t}\left(\boldsymbol{x}_i\right)^2}{\sum_{i',t'} h_{i',t'}\left(\boldsymbol{x}_{i'} \right)^2}\right\} \Mean \left[ \left\{\widehat{\phi}(\widehat{O}_{i,t})-\phi(\widehat{O}_{i,t})\right\}^2 \mid \boldsymbol{X}_i=\boldsymbol{x}_i\right] \stackrel{p}{\rightarrow} 0
\end{equation*}
when  $ \widehat{e}(\boldsymbol{x}) - e(\boldsymbol{x} )   \stackrel{p}{\rightarrow} 0 $ for any $\boldsymbol{x}$.

\textbf{Step 4: Summarize and prove Theorem \ref{thm:H2SDRL}.}
In Step 1 and Step 3, we have proved that
\begin{itemize}
	\item The regression estimator $\widehat{\mathbb{E}}_n\{\widehat{\phi}(\widehat{O}_{i,t}) \mid \boldsymbol{X}=\boldsymbol{x}\}$ is stable with respect to distance \begin{equation*} d(\widehat{\phi}, \phi)=\|\widehat{\phi}-\phi\|_{h^2} \equiv \sum_{i,t}\left\{\frac{h_{i,t}\left(\boldsymbol{x}_i\right)^2}{\sum_{i',t'} h_{i',t'}\left(\boldsymbol{x}_{i'} \right)^2}\right\} \int\{\widehat{\phi}(\widehat{O}_{i,t})-\phi(\widehat{O}_{i,t})\}^2 d \mathbb{P}\left(\widehat{O}_{i,t} \mid \boldsymbol{X}_i=\boldsymbol{x}_i\right) \end{equation*}  in the sense of Definition 1 in \cite{kennedy2020optimal}. 
	\item $d(\widehat{\phi}, \phi) \stackrel{p}{\rightarrow} 0$ when  $ \widehat{e}(\boldsymbol{x}_i) - e(\boldsymbol{x}_i )   \stackrel{p}{\rightarrow} 0 $ for any $\boldsymbol{x}_i$.
\end{itemize}
Thus, by Proposition 1 in \cite{kennedy2020optimal}, we have that
\begin{equation*}
\widehat{\tau}_{DR}(\boldsymbol{x})-\widehat{\mathbb{E}}\{\phi(\widehat{O}_{i,t}) \mid \boldsymbol{X}_i = \boldsymbol{x}\}
=\widehat{\mathbb{E}}_n\{\widehat{b}(\boldsymbol{X}_i) \mid \boldsymbol{X}_i=\boldsymbol{x}\}+o_{\mathbb{P}}\left(R_n^*(\boldsymbol{x})\right),
\end{equation*}
with $R_n^*(\boldsymbol{x})^2=\mathbb{E}\left[\widehat{\mathbb{E}}\{\phi(\widehat{O}_{i,t}) \mid \boldsymbol{X}_i = \boldsymbol{x}\} - \mathbb{E}\{\phi(\widehat{O}_{i,t}) \mid \boldsymbol{X}_i = \boldsymbol{x}\} \right]^2$ and 
\begin{equation*}
\widehat{b}(\boldsymbol{x})= \mathbb{E}\left\{\widehat{\phi}(\widehat{O}_{i,t})-\phi(\widehat{O}_{i,t}) \mid  \boldsymbol{X}_i=\boldsymbol{x}\right\} = \frac{1}{2}   \left\{     \frac{1}{1- \widehat{e}(\boldsymbol{x})} -  \frac{1}{ \widehat{e}(\boldsymbol{x})}\right\}   \left\{ \widehat{e}(\boldsymbol{x})-e(\boldsymbol{x}) \right\}\left\{ \tilde{\tau}(\boldsymbol{x})   - \widehat{\tau}(\boldsymbol{x}) \right\},
\end{equation*}
which is calculated in Step 2.

\subsection{Proof for Corollary \ref{coro:dr}}
Since the treatment effect is linear, $\tau(\boldsymbol{x})=\boldsymbol{x}' \beta$ and $\widehat{\mathbb{E}}_n$  is a linear regression, we have
\begin{equation*}
\hat{\beta}=\left(\sum_{t=T_0+1}^{T}  \sum_{i=1}^{N} \boldsymbol{x}_{i}\boldsymbol{x}_{i}^{\prime} \right)^{-1} \sum_{t=T_0+1}^{T} \sum_{i=1}^{N} \boldsymbol{x}_{i} \tilde{\Delta}_{i,t}^{D_{i,t}}=\sum_{t=T_0+1}^{T} \sum_{i=1}^{N}  \left\{ \left(\sum_{t=T_0+1}^{T}  \sum_{i=1}^{N} \boldsymbol{x}_{i}\boldsymbol{x}_{i}^{\prime} \right)^{-1} \boldsymbol{x}_{i}  \right\}\tilde{\Delta}_{i,t}^{D_{i,t}},
\end{equation*}
and
\begin{equation*}
R_n^*(\boldsymbol{x})^2=\mathbb{E}\left[\widehat{\mathbb{E}}\{\phi(\widehat{O}_{i,t}) \mid \boldsymbol{X}_i = \boldsymbol{x}\} - \mathbb{E}\{\phi(\widehat{O}_{i,t}) \mid \boldsymbol{X}_i = \boldsymbol{x}\} \right]^2= O_{\mathbb{P}} \left( (m+n)^{-1} T_1^{-1} \right).
\end{equation*} 
Note that
\begin{equation*}
\begin{aligned}
\widehat{\mathbb{E}}\{\phi(\widehat{O}_{i,t}) \mid \boldsymbol{X}_i = \boldsymbol{x}\} =\boldsymbol{x}_i^{\prime} \hat{\beta}
& =\boldsymbol{x}^{\prime} \sum_{t=T_0+1}^{T} \sum_{i=1}^{N}  \left\{ \left(\sum_{t=T_0+1}^{T}  \sum_{i=1}^{N} \boldsymbol{x}_{i}\boldsymbol{x}_{i}^{\prime} \right)^{-1} \boldsymbol{x}_{i}  \right\}\tilde{\Delta}_{i,t}^{D_{i,t}}\\
& =\boldsymbol{x}^{\prime} \sum_{t=T_0+1}^{T} \sum_{i=1}^{N}  \left\{ \left(\sum_{t=T_0+1}^{T}  \sum_{i=1}^{N} \boldsymbol{x}_{i}\boldsymbol{x}_{i}^{\prime} \right)^{-1} \boldsymbol{x}_{i}  \right\} \left( \Delta_{i,t}^{D_{i,t}} +\varepsilon_{i,t}^{1-D_{i,t}} +  e_{i,t} \right).
\end{aligned}
\end{equation*}
Thus, similarly as Lemma \ref{lemma:oneside}, we can prove that 
\begin{equation*}
\Mean \left( \left[ \widehat{\mathbb{E}}\{\phi(\widehat{O}_{i,t}) \} - \tau(\boldsymbol{X}_i) \right]^2 \mid \boldsymbol{X}_i = \boldsymbol{x}\right)  = O_{\mathbb{P}} \left( \left\{(m+n)^{-1}   +T_0 ^{-1}\right\}T_1^{-1}\right),
\end{equation*}
i.e.,
\begin{equation*}
\Mean \left[ \widehat{\mathbb{E}}\{\phi(\widehat{O}_{i,t}) \} - \tau(\boldsymbol{X}_i)  \mid \boldsymbol{X}_i = \boldsymbol{x}\right]  = O_{\mathbb{P}} \left( T_1^{-1/2} \sqrt{(m+n)^{-1}   +T_0 ^{-1}}\right),
\end{equation*}
Since the propensity score $\widehat{e}(\cdot)$  is consistently estimated  and $ |\widehat{e}(\cdot)-e(\cdot)| |\widehat{\tau}(\cdot)-\tilde{\tau}(\cdot)|= O_{\mathbb{P}} \left( (m+n)^{-1/2}T_1^{-1/2} \right)$, we have
\begin{equation*}
\widehat{b}(\boldsymbol{x})  = \frac{1}{2}   \left\{     \frac{1}{1- \widehat{e}(\boldsymbol{x}_i)} -  \frac{1}{ \widehat{e}(\boldsymbol{x}_i)}\right\}   \left\{ \widehat{e}(\boldsymbol{x}_i)-e(\boldsymbol{x}_i) \right\}\left\{ \tilde{\tau}(\boldsymbol{x}_i)   - \widehat{\tau}(\boldsymbol{x}_i) \right\} =O_{\mathbb{P}} \left( (m+n)^{-1/2}  T_1^{-1/2}\right).
\end{equation*}
Hence, with probability $1-o(1)$
\begin{equation*}
\begin{aligned}
\widehat{\tau}_{DR}(\mathbf{x}) - \tau(\mathbf{x})  & = 
\widehat{\tau}_{DR}(\mathbf{x})-\widehat{\mathbb{E}}\{\phi(\widehat{O}_{i,t}) \mid \boldsymbol{X}_i = \boldsymbol{x}_i\}+\widehat{\mathbb{E}}\{\phi(\widehat{O}_{i,t}) \mid \boldsymbol{X}_i = \boldsymbol{x}_i\} - \tau(\mathbf{x}) \\
& = \widehat{\mathbb{E}}_n\{\widehat{b}(\boldsymbol{X}) \mid \boldsymbol{X}=\boldsymbol{x}\}+o_{\mathbb{P}}\left(R_n^*(\boldsymbol{x})\right) +\widehat{\mathbb{E}}\{\phi(\widehat{O}_{i,t}) \mid \boldsymbol{X}_i = \boldsymbol{x}_i\} - \tau(\mathbf{x}) \\
& = O_{\mathbb{P}} \left( (m+n)^{-1/2} T_1^{-1/2} \right)  +o_{\mathbb{P}} \left( (m+n)^{-1/2}  T_1^{-1/2}\right) + O_{\mathbb{P}} \left( T_1^{-1/2} \sqrt{(m+n)^{-1}   +T_0 ^{-1}}\right)\\
& =O_{\mathbb{P}} \left( T_1^{-1/2} \sqrt{(m+n)^{-1}   +T_0 ^{-1}}\right). 
\end{aligned}
\end{equation*}
Therefore, we have
\begin{equation*}
\mathbb{E} \left[ \left\{\widehat{\tau}_{DR}(\mathbf{x}) - \tau(\mathbf{x})\right\}^2 \mid \boldsymbol{x}  \right]  =   O_{\mathbb{P}} \left( \left\{(m+n)^{-1}   +T_0 ^{-1}\right\}T_1^{-1}\right).
\end{equation*}

\end{document}